\documentclass[sigconf]{acmart}

\usepackage{makecell}
\usepackage{amsmath,amssymb,amsfonts}
\usepackage{amsthm}
\usepackage{multirow}
\usepackage{caption}
\usepackage{subcaption}
\usepackage{color} 
\usepackage{colortbl}
\usepackage{multirow}

\usepackage{lipsum}
\usepackage{enumitem}
\usepackage{graphicx}
\usepackage{lipsum}

\usepackage[framemethod=tikz]{mdframed}
\definecolor{mygray}{RGB}{243,243,244}
\newmdenv[
  innertopmargin=0pt,
  backgroundcolor=mygray,
  linecolor=none,
  innerleftmargin=0pt,
  innerrightmargin=0pt,
  leftmargin=0pt
  ]{mymath}

\usepackage[ruled, linesnumbered,vlined]{algorithm2e}

\AtBeginDocument{%
  }

\copyrightyear{2025} 
\acmYear{2025} 
\setcopyright{acmlicensed}\acmConference[WWW '25]{Proceedings of the ACM Web Conference 2025}{April 28-May 2, 2025}{Sydney, NSW, Australia}
\acmBooktitle{Proceedings of the ACM Web Conference 2025 (WWW '25), April 28-May 2, 2025, Sydney, NSW, Australia}
\acmDOI{10.1145/3696410.3714952}
\acmISBN{979-8-4007-1274-6/25/04}

\begin{document}

% \title{Language-like or Structural: Learning a Graph Foundation Model from Structural Geometry}
\title{RiemannGFM: Learning a Graph Foundation Model from Riemannian Geometry}

\author{Li Sun}
\authornote{Corresponding Author: Li Sun, ccesunli@ncepu.edu.cn.}
\email{ccesunli@ncepu.edu.cn}
\affiliation{%
  \institution{North China Electric Power University}
  \city{Beijing}
  \country{China}
  \postcode{102206}
}

\author{Zhenhao Huang}
\email{huangzhenhao@ncepu.edu.cn}
\affiliation{%
  \institution{North China Electric Power University}
  \city{Beijing}
  \country{China}
  \postcode{102206}
}

\author{Suyang Zhou}
\email{zhousuyang@ncepu.edu.cn}
\affiliation{%
  \institution{North China Electric Power University}
  \city{Beijing}
  \country{China}
  \postcode{102206}
}

\author{Qiqi Wan}
\email{wanqiqi@ncepu.edu.cn}
\affiliation{%
  \institution{North China Electric Power University}
  \city{Beijing}
  \country{China}
  \postcode{102206}
}

\author{Hao Peng}
\email{penghao@buaa.edu.cn}
\affiliation{%
  \institution{Beihang University}
    \city{Beijing}
  \country{China}
}

\author{Philip  Yu}
\email{psyu@uic.edu}
\affiliation{%
  \institution{University of Illinois, Chicago}
  \country{USA}
}

% \affiliation{%

\renewcommand{\shortauthors}{Li Sun et. al.}

\begin{abstract}
% and has achieved tremendous success in natural language processing, e.g., GPT-4, as well as computer vision. 

% but the foundation model encounters significant challenges in the graph domain due to its non-Euclidean nature.
% So far, existing graph foundation models are built upon the Large Language Model.
% % , which make efforts to align graph structures with the language and to seek advanced prompting techniques.
% % Consequently, the universality and transferability heavily 
% They either rely on the lingual vocabulary over textual features, 
% limiting the usage to a wide range of graphs other than the text-attributed ones, 
% or make efforts to align graph structures with the language and to seek advanced prompting techniques.
% Though this line of studies receive encouraging performance, 

%pretraining a single, universal model generalizable to different downstream tasks.
The foundation model has heralded a new era in artificial intelligence, 
pretraining a single model to offer cross-domain transferability on different datasets.
% Graphs are 
% , ranging from recommender systems to biochemical structures.
Graph neural networks excel at learning graph data, the omnipresent non-Euclidean structure, but often lack the generalization capacity.
Hence, \textbf{graph foundation model} is drawing increasing attention, and 
recent efforts have been made to leverage Large Language Models.
%, encouraged by the remarkable success of GPT-4.
On the one hand, existing studies primarily focus on text-attributed graphs, while a wider range of real graphs do not contain fruitful textual attributes.
On the other hand,  the sequential graph description tailored for the Large Language Model neglects the structural complexity, which is a predominant characteristic of the graph.
Such limitations motivate an important question: \textbf{Can we go beyond Large Language Models, and pretrain a universal model to learn the structural knowledge for any graph?}
The answer in the language or vision domain is a shared vocabulary.
We observe the fact that there also exist shared substructures underlying graph domain, 
and thereby open a new opportunity of graph foundation model with structural vocabulary.
% (by which any graph can be constructed).
The key innovation is the discovery of a simple yet effective structural vocabulary of trees and cycles, and we explore its inherent connection to Riemannian geometry.
Herein,  we present a universal pretraining model, \texttt{RiemannGFM}.
%, with geometric contrastive learning.
Concretely, we first construct a novel product bundle to incorporate the diverse geometries of the vocabulary. 
Then, on this constructed space, we stack Riemannian layers where the structural vocabulary, regardless of specific graph, is learned in Riemannian manifold offering  cross-domain transferability.
%and node encoding is generated in the tangent space for arbitrary input graph.
Extensive experiments show the effectiveness of \texttt{RiemannGFM} on a diversity of real graphs.
% in a novel augmented product manifold where each node has the coordinate in manifold factor, and is attached to node encoding in tangent bundle factor.
% We design the vocabulary learning module with cross-geometric attention to study how to place a tree (cycle) in , and global learning module to learn the structural knowledge for any graph.
% In other words, existing solutions neglect the fruitful information on the graph structure itself, which is the predominant character on graphs.
% A natural question arises that 
% \textbf{can we go beyond the Large Language Model, and pretrain a universal model to learn the structural knowledge for any graph?}
% In this paper, we approach this problem  by learning a graph foundation model from structural geometry.
% %, grounded by the solid theory of .
% Thus, we construct the graph foundation model on the product of Rimannian mainfolds
% (named \textbf{\texttt{RiemannGFM}})\footnote{Codes are available at ......}.
% Specifically, ..... grounded  on the theory of Riemannian harmonic analysis. 
\vspace{-0.1in}
\end{abstract}

%\footnote{Codes are available at ......} 







\maketitle

%!TEX root = ./main.tex

% \newpage
\section{Introduction}

%Graph Neural Networks (GNNs) [] excel at graph representation learning 

Designing a foundation model has been a longstanding objective in artificial intelligence that pre-trains a single, universal model on massive data allowing for cross-domain transferability on different datasets. 
%so as to support a wide range of downstream tasks.
Recently, the Large Language Model (LLM) such as GPT-4 \cite{2024gpt4} marks a revolutionary advancement of the foundation model in the language realm.
In the real world, graphs are also ubiquitous, describing the data from Web applications, social networks, biochemical structures, etc.
Unlike word sequences in the language, graphs present distinct, non-Euclidean structures encapsulating the complex intercorrelation among objects,
which prevents the direct deployment of LLM.
%In the graph domain, 
Graph Neural Networks (GNNs) \cite{nips18hnn,iclr17gcn,iclr18gat,icml19sgc} conduct neighborhood aggregation over the graph and achieve state-of-the-art performance on learning graph data.
The significant limitation of GNNs is the lack of generalization capacity. 
GNNs are often designed for specific tasks, and re-training is typically required on different tasks or datasets to maintain expressiveness.
Consequently, \textbf{Graph Foundation Models} (GFMs) are emerging as an interesting research topic in the graph domain.
%Encouraged by the remarkable success in language domain , 
% Compared to the significant success in the language, the development of GFMs is still in the infant stage. 
% Existing solutions typically follow the paradigm of Large Language Model.

%Compared to the tremendous success of the foundation model in other domains, GFM is still in the infant stage.
Pioneering work \cite{kdd23all-in-one} designs the graph prompting to unify the downstream tasks, the analogy to language prompt.
GCOPE \cite{kdd24gcope} further conducts model training on multi-domain graphs with coordinators in order to improve the generalization capacity to different datasets.
Recent efforts have been made to integrate GNN with LLM. 
For example, LLaGA \cite{icml24llaga} develops a graph translation technique that reshapes a graph into node sequences,
while OFA \cite{iclr24ofa}  unifies different graph data by describing nodes and edges with natural language.
Also, there are successful practices in specific domains (e.g., knowledge graphs \cite{nips23prodigy,iclr24ultra} and molecular structures \cite{iclr24Beaini}) or specific tasks (e.g., node classification \cite{zhao2024graphany}), which are far from the universal GFM.

On the one hand, existing studies primarily focus on the text-attributed graphs.
The structural knowledge is coupled with textual attributes (or language description), and the transferability relies on the commonness of text \cite{icml24position}. 
Consequently, it leads to suboptimal performance on the graphs other than text-attributed ones, as investigated in our Experiment as well.
However,  there exists a wide range of graphs that do not contain fruitful textual attributes, and may only have structural information.
An alternative perspective is to seek the transferability from the structures  (e.g., the common substructures), so that such knowledge is applicable to any graph.
%Their generalization capacity generally relies on the vocabulary over textual features inherited from LLM.
Surprisingly, it has not yet been touched in the context of GFM.

On the other hand, the sequential graph description, tailored for the language model, tends to fail in capturing the structural complexity, which is a predominant characteristic of graphs.
GFMs so far work with the traditional Euclidean space, while recent advances report superior expressiveness of hyperbolic spaces in learning tree-like (hierarchical) graphs \cite{icml18HierarchiesofHyperbolic,nips19hgcn}.
%Existing GFMs , trivializing the structural geometry of real graphs.
However, modeling the structural complexity is challenging.
For instance, 
hyperbolic models trained on tree-like graphs cannot be generalized to those of different structures.
In addition, graph structures are indeed quite complex, tree-like in some regions and cyclical in others \cite{iclr19mixCurvature}.
%graph structures are rather complex indeed, in some regions tree-like, while in others cyclical .
% We also notice the product manifold is leveraged to fit the geometry of specific graphs \cite{iclr23lantentGraphProduct,aaai22selfMG}, 
We also notice the product manifold is leveraged to fine-tune  the geometry of given structures \cite{iclr23lantentGraphProduct,aaai22selfMG},  
which is orthogonal to our focus.
In other words,  it is still not clear how to connect such geometric expressiveness to GFM. 
%, while the hallucination inherited from LLM may generate incorrect answer on graphs as well.

%A natural question arises that 
Motivated by the aforementioned limitations, we raise an important question: 
\textbf{Can we go beyond Large Language Models, and pre-train a universal model to learn the structural knowledge for any graph?}
The answer to the foundation model in language or vision domain is a shared vocabulary \cite{chang2024surveyllm}.
In fact, there also exist common substructures underlying the graph domain, 
and the observation offers us a fresh perspective to build graph foundation models.
Accordingly, we introduce the concept of \textbf{structural vocabulary} by which any graph can be constructed.
%which is a collection of substructures that are able to construct any graph.
The key innovation of this paper is the discovery of  a simple yet effective structural vocabulary
consisting of substructures of trees and cycles (e.g., node triangles). 
We  explore the inherent connection between the structural vocabulary and \textbf{Riemannian geometry},
where hyperbolic space aligns  tree structures \cite{nips18hnn,2012lowDistort}, while hyperspherical space is suitable to cycles \cite{iclr19mixCurvature,Petersen16}.

Accordingly, it calls for a representation space to model both local geometry (trees or cycles) and graph structure.
To this end, we for the first time introduce the tangent bundle to the graph domain, coupling a  Riemannian manifold and its surrounding tangent spaces.
We leverage the node coordinate on the manifold to embed the local geometry,
while the node encoding in tangent spaces accommodates the information of graph structure.
Grounded on the elegant framework of Riemannian geometry, we present a universal pre-training model (\textbf{\texttt{RiemannGFM}})  on the product bundle to incorporate the vocabulary of diverse geometries.
\texttt{RiemannGFM} stacks the universal Riemannian layer, which consists  of a vocabulary learning module and a global learning module.
In the vocabulary learning module, we focus on embedding the structural vocabulary into Riemannian manifolds, regardless of the specific graph.
Specifically, this involves updating the node coordinates of a tree (or cycle) in hyperbolic (or hyperspherical) space.
For each substructure, cross-geometry attention is formulated in the manifolds in which we derive a manifold-preserving linear operation.
The global learning module is responsible for updating the node encoding.
With multiple substructures sampled in the graph, we first perform substructure-level aggregation by the proposed bundle convolution, solving the incompatibility issue over tangent bundle, 
and then calculate the graph-level node encoding with the geometric midpoint and parallel transport.
Finally, we conduct geometric contrastive learning among different views, provided by different geometries, so that \texttt{RiemannGFM}  is capable of generating informative node encoding for an arbitrary graph, 
underpinned by the shared structural knowledge learned in Riemannian geometry.

\textbf{Contribution Highlights.} 
Overall, key contributions are three-fold:
\textbf{A. Foundation Model for Graph Structures.} 
We explore GFM  for a wider range of real graphs, not limited to text-attributed ones, 
and for the first time study GFM from structural geometry to the our best  knowledge.
\textbf{B. Universal Riemannian Pre-training.} 
We propose a universal pre-trained model (\texttt{RiemannGFM}) on a novel product bundle
where  the structural vocabulary is learned in Riemannian manifold, offering the shared structural knowledge for cross-domain transferability.
%the normalized Fourier encoding offers the flexibility to handle different graphs, and
% tuee from the structural geometry ...
\textbf{C. Extensive Experiments.} 
We evaluate the superiority of \texttt{RiemannGFM} in cross-domain transfer learning and few-shot learning on a diversity of real graphs.
%, and discuss the effectiveness of structural vocabulary.

\begin{figure*}
\centering
    \includegraphics[width=0.93\linewidth]{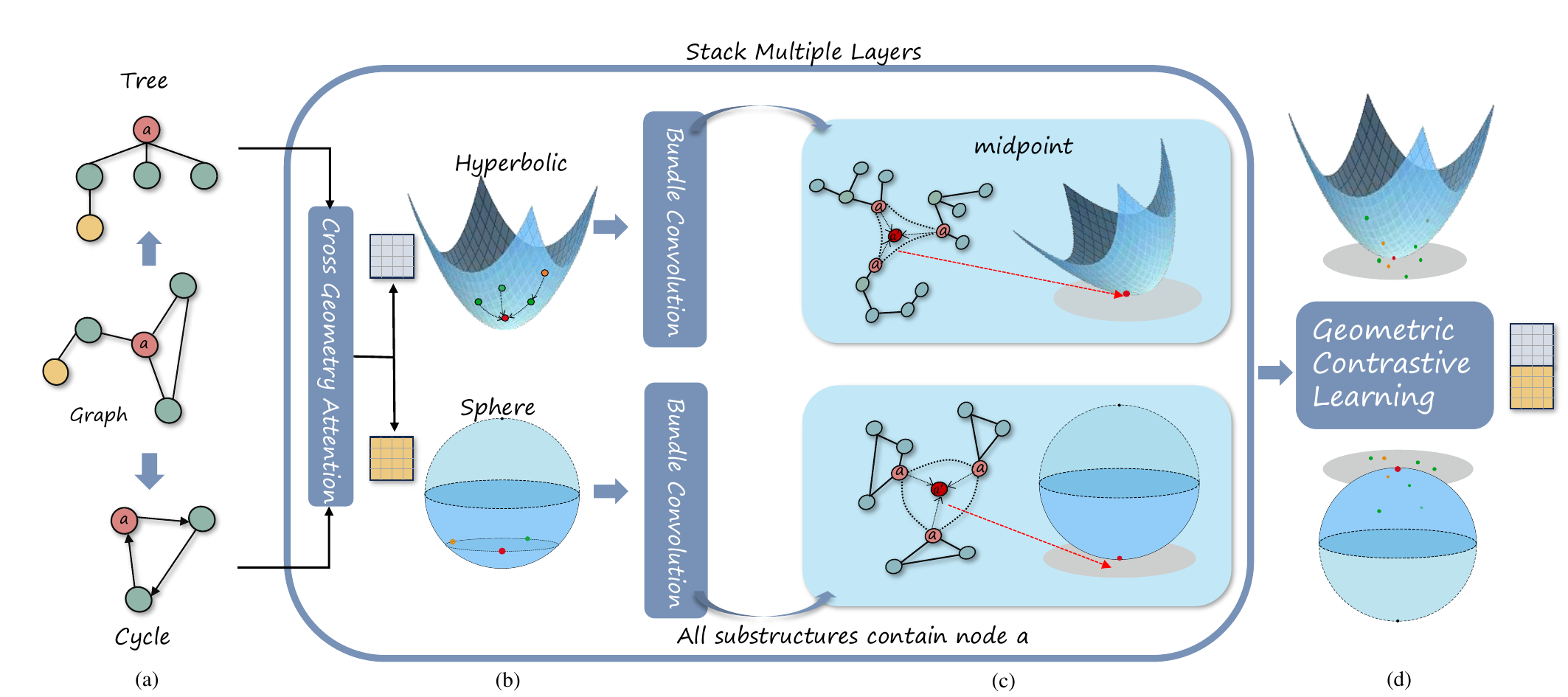}
    \vspace{-0.15in}
        \caption{Overall architecture of the proposed graph foundation model: \textbf{\texttt{RiemannGFM}}.}
    \label{fig:overall}
        \vspace{-0.15in}
\end{figure*}

\vspace{-0.1in}
\section{Preliminaries}
% This section provides an introduction to the basic concepts that underpin our proposed framework. We first formally review the concepts of Riemannian Geometry and the Lorentz/Spherical Model, then formulate a novel problem of graph foundation model from structural geometry. Important notations are summarized in Appendix A.

% This section reviews the basic concepts of Riemannian Geometry and the model space of Lorentz/Spherical Model, and then formally describes the novel problem of graph foundation model from structural geometry. Important notations are summarized in Appx. A.

\vspace{-0.05in}
\subsubsection*{\textbf{Riemannian Geometry.}}
Geometrically, a complex structure is related to a Riemannian manifold, which is a smooth manifold $\mathcal M$ endowed with a Riemannian metric $\mathfrak{g}$.
%Riemannian manifold is the basic objection of Riemannian geometry, which is a smooth manifold M endowed with a Riemannian metric g. 
Each point $\boldsymbol x$ in the manifold is associated with a tangent space $\mathcal T_x \mathcal M$ where the metric $\mathfrak{g}$ is defined. 
The mapping between the tangent space and manifold is done via exponential and logarithmic maps,
and parallel transport conducts the transformation between two tangent spaces.
The geodesic between two points is the curve of the minimal length that connects them on the manifold.
%The logarithmic map transforms the vector in the manifold to the tangent space, while the exponential map doer the inverse transform. Euclidean space is a special case of Riemannian manifold. 
The curvature $\kappa_x$ is the geometric quantity measuring the extent of how a surface deviates from being flat at $\boldsymbol x$. 
A manifold is referred to as a \textbf{Constant Curvature Space} (CCS) if and only if curvature $\kappa_x$ is equal everywhere, so that the closed-form metric is derived.
There exist three types of CCS (a.k.a. isotropic manifold): hyperbolic space $\mathcal H$ with negative curvature, hyperspherical space $\mathcal S$ with positive curvature, and zero-curvature Euclidean space,  a special case of Riemannian geometry. 

%$\mathcal{L}^d_\kappa=\left\{x\in \mathbb{R}^{n+1} \mid \langle x, x \rangle_{\mathcal{L}}=\frac{1}{K}, x_{t}>0\right\}$,
\vspace{-0.05in}
\subsubsection*{\textbf{Lorentz/Spherical Model.}}
Here, we give a unified formalism of hyperbolic and hyperspherical space.
Specifically, a $d$-dimensional CCS with constant curvature $\kappa$ ($\kappa \neq 0$) is defined on the smooth manifold of 
$\mathcal L_{\kappa }^{d}=\{ \boldsymbol x=
\left[\begin{array}{c}
x_t  \\
\boldsymbol x_s 
\end{array}\right] \in \mathbb R^{d+1} 
| \langle \boldsymbol x, \boldsymbol x \rangle_\kappa = \frac{1}{\kappa}, x_t >0, \boldsymbol x_s \in \mathbb R^d\}$
equipped with the curvature-aware inner product as follows,
\vspace{-0.05in}
\begin{equation}
    \langle \boldsymbol x, \boldsymbol y \rangle_\kappa:=\operatorname{sgn}(\kappa) x_{t} y_{t}+\boldsymbol x_{s}^{\top}\boldsymbol y_{s}, \quad  \boldsymbol x, \boldsymbol y \in \mathcal L_{\kappa }^{d},
    \vspace{-0.03in}
    \label{manifold}
\end{equation}
where
% $x_t$ and $\boldsymbol x_s$ denote the time-dimension and space-dimension, respectively.
    $\operatorname{sgn}$ is the sign function and thereby the Riemannian metric at $\boldsymbol x$ is induced as 
$\mathfrak{g}_x= \operatorname{diag}(\operatorname{sgn}(k), 1, \ldots, 1)$, a diagonal matrix.
The north pole of $\mathcal L_{\kappa }^{d}$ is given as $\boldsymbol o=[\frac{1}{\sqrt{|\kappa|}}, 0, \cdots, 0]^\top$.
Closed-form exponential and logarithmic maps exist (detailed in Appendix D).

\vspace{-0.05in}
\subsubsection*{\textbf{Notations \& Problem Formulation.}}
A graph $\mathcal G$ is defined over node set $\mathcal V$ and edge set $\mathcal E \subseteq \mathcal V \times \mathcal V$, 
and each node is optionally associated with an attribute $\boldsymbol x$.
Note that, the attribute is not necessarily required given a wide range of non-attributed graphs exist in the real world.
Analogy to the foundation model for the language, the graph foundation model aims to pre-train a single, universal model $\mathbf \Phi$ parameterized by $\mathbf \Theta$, which is applicable to other graphs to generate informative representations $\boldsymbol z$ for downstream tasks (e.g., node-level and edge-level).
In particular, the model $\mathbf \Phi$ offers the \textbf{cross-domain transferability} that the parameters $\mathbf \Theta$ pre-trained on one domain can be utilized on another domain with slight treatments.
In this paper, we argue that GFM should also offer the \textbf{universality} to any graph (not limited to textual-attributed graphs), and highlight the inherent \textbf{structural geometry} which is largely ignored in the previous GFMs.

\section{\textbf{\texttt{RiemannGFM}}: Learning Structural Vocabulary in Riemannian Geometry}

\vspace{-0.02in}
Different from previous GFMs coupling the structural knowledge with textual attributes,
%mimicking or aligning to the Large Language Model,
we put forward a fresh perspective of studying graph structures, and propose a universal pre-training model named \texttt{RiemannGFM},
which is capable of learning the structural knowledge so as to offer the cross-domain transferability among a wider range of real graphs.
The key novelty lies in that we discover an effective \emph{structural vocabulary} for any graph structure and explore its connection to \emph{Riemannian Geometry}.
At the beginning, we introduce a novel concept of  structural vocabulary.
%\vspace{-0.05in}
 \begin{mymath}
\newtheorem*{def1}{Definition 1 (Structural Vocabulary)} 
\begin{def1}
A collection of substructures is said to be a structural vocabulary when they are able to construct an arbitrary graph.
\end{def1}
\end{mymath}
\vspace{-0.1in}
\subsubsection*{\textbf{Structural Vocabulary and Constant Curvature Spaces (CCS)}}
%The structural vocabulary is a collection of graph substructures by which any graph can be constructed.
The answer to the foundation model in language or vision domain is a shared vocabulary.
For a language model, the text is broken down into smaller units such as words by which the commonness and transferability are encoded \cite{chang2024surveyllm}.
Analogous to word vocabulary of the language, the structural vocabulary considers the shared units in graph domain.
In \texttt{RiemannGFM}, we leverage a simple yet effective structural vocabulary, consisting of trees and cycles.
An intuitive example is that a tree and cycles with coinciding edges form an arbitrary connected component unless it is exactly a tree.
On modeling the vocabulary, we notice the fact that a tree cannot be embedded in Euclidean space with bounded distortion\footnote{Embedding distortion is measured by $\frac{1}{|\mathcal V|^2} \sum_{ij}\left| \frac{d_G(v_i, v_j)}{d(\mathbf x_i, \mathbf x_j)}-1\right|$, where each node $v_i \in \mathcal V$ is embedded as $\mathbf x_i$ in representation space. $d_G$ and $d$ denote the distance in the graph and the space, respectively.},
while the embedding distortion is proved to be bounded in low-dimensional hyperbolic spaces \cite{2012lowDistort}.
The geometric analogy of cycle is the hyperspherical space, as both cycle and hyperspherical space present the rotational invariant \cite{Petersen16}.
Therefore, we propose to utilize the constant curvature spaces (i.e., hyperbolic and hyperspherical spaces) to model the substructures of trees and cycles.
% Note that there exists no isometric mapping between CCS and Euclidean one, and we discuss the significance of introducing CCS in the Experiment.
%In this paper, we opt for the Lorentz/Spherical model in Eq. (\ref{manifold}) of CCS.
% owing to its numerical stability.

\vspace{-0.07in}
\subsubsection*{\textbf{Overall Architecture.}}
We design the universal pre-training model (\textbf{\texttt{RiemannGFM}}) on the product bundle in light of the diverse substructures in the vocabulary.
The overall architecture is illustrated in Fig. 1.
%\texttt{RiemannGFM} first performs graph normalizing with the normalized Fourier encoding (\textbf{Sec. \ref{sec:nfe}}).
According to the structural vocabulary, we first sample the trees and cycles in the graph, as shown in Fig. 1(a).
Subsequently, we stack the universal Riemannian layers  on the product bundle, consisting of the Vocabulary Learning Module in Fig. 1(b) and Global Learning Module in Fig. 1(c).
Without graph augmentation, \texttt{RiemannGFM} is pre-trained with the geometric contrastive loss in Fig. 1(d).

% \subsection{Normalized Fourier Encoding} \label{sec:nfe}
% Graph normalizing is challenging. ..
% a novel non-parametric formulation, normalized Fourier encoding, ...
% whose another advantage is that the output encoding already resides in the corresponding CCS.

% a predefined number of  harmonic waves.

% \newtheorem*{pro1}{Proposition 1 (Isometric Invariant)} 
% \begin{pro1}
% Isometric Invariant
% \end{pro1}
% \begin{proof}
% The proof is detailed in Appendix B.
% \end{proof}

% \subsubsection*{\textbf{Remark.}}

% \subsubsection{\textbf{Manifold-preserving Transformation}}

% \newtheorem*{pro2}{Proposition 2 (Manifold Preserving)} 
% \begin{pro2}
% Manifold Preserving
% \end{pro2}
% \begin{proof}
% We sketch the proof with key ideas, and present the details in the Appendix B.
% \end{proof}

\vspace{-0.1in}
\subsection{Universal Riemannian Layer}  \label{sec:url}
In the heart of \texttt{RiemannGFM}, we stack the universal Riemannian layer on the product bundle, where
Vocabulary Learning Module  performs cross-geometry attention to \textbf{embed the structural vocabulary into CCSs, regardless of specific graphs, thus offering the shared structural knowledge for cross-domain transferability}.
Global Learning Module  aligns different substructures with a global view and, accordingly, generates node encodings through the proposed bundle convolution.
% through a hierarchical aggregation mechanism.

% On this product manifold, we stack the Riemannian layers where the  is learned in Riemannian manifold, 
% offering the shared structural knowledge for cross-domain transferability, 
% and informative node encoding is generated in the tangent bundle for arbitrary input graph. 

% The transferability lies in the structural vocabulary underlying graph domain.

\vspace{-0.05in}
\subsubsection{\textbf{A Layer on the Product Bundle}}
We elaborate on a novel representation space for GFM, where the \textbf{tangent bundle} is introduced to graph domain for the first time.
In Riemannian geometry, a tangent bundle\footnote{A tangent bundle is defined as a smooth manifold $\mathcal M$ attached with a disjoint union of  tangent spaces surrounding it 
 $\mathcal T\mathcal M=\bigsqcup_{\boldsymbol x \in \mathcal M}\mathcal T_{x}\mathcal M$.} typically consists of  (1) a CCS highlighting the local geometry, and (2) the tangent spaces describing the complementary information \cite{Petersen16}. 
 In our design, each node $i$ in the bundle is associated with node coordinate and node encoding.
Concretely,
 the  coordinate  in the manifold $\boldsymbol p_i \in \mathcal M$ contains the relative position in substructures (i.e., structural vocabulary),
 while the  encoding in tangent space $\boldsymbol z_i \in\mathcal T_{\boldsymbol p_i}\mathcal M$ carries the information of global structure (in the graph level).
To incorporate the substructures of different geometries, we construct a \textbf{product bundle} as
    \vspace{-0.05in}
\begin{equation}
\mathcal P^{d_P}=\left(\mathcal H^{d_H}_{\kappa_H} \otimes \mathcal T\mathcal H^{d_H}_{\kappa_H}\right)\otimes \left(\mathcal S^{d_S}_{\kappa_S} \otimes \mathcal T\mathcal S^{d_S}_{\kappa_S}\right) ,  d_P=2d_H+2d_S,
\label{eq:space}
\end{equation}
where  $\otimes$ denotes Cartesian product, $d_{(\cdot)}$ and $\kappa_{(\cdot)}$ are the dimension and curvature, respectively.
%In particular, 
For each node in this product, we have 
$\boldsymbol x_i =[\boldsymbol p^H_i||\boldsymbol z^H_i||\boldsymbol p^S_i||\boldsymbol z^S_i] \in \mathcal P^{d_P}$, 
where $||$ is vector concatenation, and 
$\boldsymbol p^H_i \in \mathcal H^{d_H}_{\kappa_H}$, $\boldsymbol z^H_i \in \mathcal T_{\boldsymbol p^H_i}\mathcal H^{d_H}_{\kappa_H}$, $\boldsymbol p^S_i \in \mathcal S^{d_S}_{\kappa_S}$, $\boldsymbol z^S_i \in \mathcal T_{\boldsymbol p^S_i}\mathcal S^{d_S}_{\kappa_S}$.
Accordingly,  Riemannian metric of the product bundle is yielded as 
$\mathfrak{g}^{\mathcal P}_x=\mathfrak{g}^{\kappa_H}_x \oplus \mathbf I_{d_H+1}\oplus  \mathfrak{g}^{\kappa_S}_x \oplus  \mathbf I_{d_H+1}$, 
where $\mathbf I_{d_H+1}$ is the $(d_H+1)$-dimensional identity matrix, 
and  $\oplus$ denotes the direct sum among matrices.
In Eq. (\ref{eq:space}), hyperbolic bundle $\mathcal H^{d_H}_{\kappa_H} \otimes \mathcal T\mathcal H^{d_H}_{\kappa_H}$  and hyperspherical bundle $\mathcal S^{d_S}_{\kappa_S} \otimes \mathcal T\mathcal S^{d_S}_{\kappa_S}$ are responsible for trees and cycles, respectively. 
Our framework is applicable to multiple bundles with any curvatures, and we use the two-bundle product for simplicity.
In this paper, we opt for the unified formalism $\mathcal L^d_\kappa$ in Eq. (\ref{manifold}) for hyperbolic and hyperspherical spaces.
%$\mathcal T\mathcal H^{d_H}_{\kappa_H}$ and $\mathcal T\mathcal S^{d_S}_{\kappa_S}$ are tangent bundles attached to the manifolds.
%is hyperbolic space to study trees,   is the hyperspherical space for cycles, and $\mathcal H^{d_H}_{\kappa_H}$

% In our design, each node in the tree/cycle has a \textbf{coordinate in the factor manifold}, 
% and is associated with a \textbf{node encoding}, not necessarily adherence to manifold definition, placed \textbf{in the tangent space} of that coordinate.
% Thus, the spaces of node encoding span the corresponding factor bundle.

\subsubsection{\textbf{Deriving Riemannian Operations}}
Before designing the neural architecture, we derive a closed-form Riemannian linear operation and introduce a geometric midpoint for mathematical preparation. (Proofs are given in Appendix B.)
In Riemannian geometry, the operation output is required to remain on the manifold, i.e., manifold preserving.
Previous works typically meet this requirement by involving an additional tangent space with the exponential/logarithmic maps \cite{nips19hgcn,nips19hgnn}.
However, the lack of isometry and possible mapping error \cite{iclr23HyLa} motivate a fully Riemannian formulation.
We formulate the linear operation with matrix-left-multiplication. 
The operation parameterized by $\boldsymbol W$ is derived as 
\begin{equation}
\forall \boldsymbol x = \left[\begin{array}{c}
x_t \\
\boldsymbol x_s
\end{array}\right] \in \mathcal L_{\kappa }^{d},  \quad
f_{\boldsymbol W}(\boldsymbol x)=
\left[\begin{array}{cc}
1 & \mathbf{0}^{\top} \\
\mathbf{0} & \alpha \boldsymbol W
\end{array}\right]
\left[\begin{array}{c}
x_t \\
\boldsymbol x_s
\end{array}\right],
\end{equation}
where re-scaling factor is defined as $\alpha=\frac{\sqrt{\kappa^{-1}-sgn(\kappa)x^2_t}}{\|\boldsymbol W\boldsymbol x_s\|^2}$ and $\| \cdot \|$ denotes the $L2$ norm.
The theoretical guarantee is given below.
 \begin{mymath}
\newtheorem*{thm1}{Theorem 1 (Manifold-preserving of Proposed Operation)} 
\begin{thm1}
Given $\boldsymbol x \in \mathcal L_{\kappa }^{d_1}$ and $\kappa \neq 0$,  
$f_{\boldsymbol W}(\boldsymbol x) \in \mathcal L_{\kappa }^{d_1}$ preserves on the manifold with  any $\boldsymbol W \in \mathbb R^{{d_1}\times{d_1}}$, and 
$f_{\boldsymbol W}(\boldsymbol x) \in \mathcal L_{\kappa }^{d_2}$ holds for any $\boldsymbol W \in \mathbb R^{{d_1}\times{d_2}}$.
\end{thm1}
\end{mymath}
\noindent We notice that \citet{acl22fullyHyper} and \citet{kdd24Hypformer} propose linear operation in fully Riemannian fashion recently, but none of them allows for operating in  any constant curvature.
The aggregation is typically given as an arithmetic mean in Euclidean spaces \cite{www21Lgcn,nips17GraphSAGE}.
With a set of points and their weights $\{\boldsymbol x_i, \nu_i\}_{i \in \Omega}$, $\boldsymbol x_i \in \mathcal L_{\kappa }^{d}$, $\nu_i \in \mathbb R$,
the arithmetic mean in CCS takes the form of 
\begin{equation}
mid_\kappa(\{\boldsymbol x_i, \nu_{i}\}_{i \in \Omega})= 
\frac{1}{\sqrt{|\kappa|} } \sum\nolimits_{i \in \Omega}\frac{\nu_{i} \boldsymbol x_i}{\left| \|\sum\nolimits_{j \in \Omega} \nu_{j} \boldsymbol x_j\|_\kappa \right|}, \ \kappa \neq 0,
\label{eq:mid}
\end{equation}
with the definition of $\|\boldsymbol x\|^2_\kappa=\langle \boldsymbol x, \boldsymbol x \rangle_\kappa$.
We demonstrate the fact that the  mean in Eq. (\ref{eq:mid}) is the geometric midpoint on the manifold.
 \begin{mymath}
\newtheorem*{thm2}{Theorem 2 (Arithmetic Mean as Geometric Midpoint)} 
\begin{thm2}
%With a set of points and their weights $\{\boldsymbol x_i, \nu_i\}_{i \in \Omega}$, $\boldsymbol x_i \in \mathcal L_{\kappa }^{d}$, $\nu_i \in \mathbb R$,
The arithmetic mean in Eq. (\ref{eq:mid}) is on the manifold $\boldsymbol c=mid_\kappa(\{\boldsymbol x_i, \nu_{i}\}_{i \in \Omega}) \in \mathcal L_{\kappa }^{d}$,
and is the geometric midpoint $\boldsymbol c=\arg \min\nolimits_{\boldsymbol c \in \mathcal L_{\kappa }^{d}} \sum\nolimits_{i \in \Omega} \nu_{i} d^2_\kappa(\boldsymbol c, \boldsymbol x_i)$ w.r.t. the squared distance $d$.
\end{thm2}
\end{mymath}

% the weighted geometric centorid over the set $\bar{\mathcal N}_i$, the neighbors of $i$ and itself, i.e.,
% $\arg \min\nolimits_{\boldsymbol h_i \in \mathcal M} \sum\nolimits_{j\in \bar{\mathcal N}_i} \nu_{ij} d^2_c(\boldsymbol h_i, \boldsymbol h_j), \forall c$ and $\nu_{ij}$ denotes the attentive weight.
% For any $c$, we derived the closed form solution $\boldsymbol h_i=AGG^c(\{\boldsymbol h_j, \nu_{ij}\}| j \in \bar{\mathcal N}_i)$,
% \vspace{-0.07in}

\subsubsection{\textbf{Vocabulary Learning Module}}
This module focuses on the substructure, with the objective of embedding the structural vocabulary into the constant curvature spaces.
% in the manifolds.update node coordinates of the substructure on the factor manifold, 
In other words, we are interested in how to place a tree (or cycle) in the hyperbolic (or hyperspherical) space. 
To this end, we propose a \textbf{Cross-geometry Attention} to learn node coordinates in the substructure.
We elaborate the formulation with the tree  in hyperbolic factor.
%In $\mathcal H^d_\kappa$, 
In hyperbolic manifold, we propose to update a tree in bottom-up fashion, and thus this problem is reduced to induce the node coordinate from its descendant nodes.
The node coordinate is given by the attentional aggregation with attentional weights as follows,
% \begin{align}
%     Q, K, V = f_Q(\mathbf{Z}^N), f_K(\mathbf{Z}^M), f_V(\mathbf{Z}^M),
% \end{align}
% \begin{align}
%     (\boldsymbol{z}^M)'_i = f_o(\operatorname{Midpoint}(\{a_{ij}, \boldsymbol v_j\};j \in \operatorname{Nei}(i))),
% \end{align}
\vspace{-0.05in}
\begin{align}
    \boldsymbol v_i=mid_\kappa(\{\boldsymbol v_j, \alpha_{ij}\}_{(i,j) \in \Omega})\in \mathcal L^d_\kappa, \ \boldsymbol v_j \in \mathcal L^d_\kappa,
       \label{eq:cross-geo}
\end{align}
\vspace{-0.1in}
\begin{align}
     \alpha_{ij}=\frac{\exp(\phi([\boldsymbol q_i || \boldsymbol k_j]))}{\sum_{(i,t) \in \Omega}\exp(\phi([\boldsymbol q_i || \boldsymbol k_t]))},
     \label{eq:att}
\vspace{-0.05in}
\end{align}
where $j$ is the descendant node of $i$, and we slightly abuse $j$ to include the coordinate information of $i$ itself.
In cross-geometry attention, the key, query and value are derived with the Riemannian linear operation $\boldsymbol k_i=f_{\boldsymbol V}(\boldsymbol p^H_i)$, $\boldsymbol q_i=f_{\boldsymbol Q}(\boldsymbol p^S_i)$ and $\boldsymbol v_i=f_{\boldsymbol V}(\boldsymbol p^H_i)$, respectively.
$\phi$ can be any function that returns a scalar.
As a result, the node coordinate $\boldsymbol p^H_i$ is updated as $\boldsymbol v_i$.
Note that, the query value is given from $\mathcal S^d_\kappa$ so as to leverage the compensatory information of the other geometry.
(Compared to performing attention in a single geometry, the superiority of our design is evaluated in the Ablation Study.)
In addition, the proposed aggregation is \textbf{unidirectional} which is different from that of traditional bidirectional aggregations in graph model \cite{iclr17gcn,nips17GraphSAGE,www21Lgcn}. 
In traditional aggregations, each node considers its information in the neighborhood, and vice versa.
However, as in Eq. (\ref{eq:cross-geo}), each node receives the coordinates of the descendant nodes to locate itself on the manifold, 
while the reverse information path does not exist, that is, the node's coordinate is not affected by the ancestor node in bottom-up construction.

Similarly, the cycle is refined on hyperspherical manifold where the node coordinate is updated by the two nodes connecting it.
The unidirectional path is from neighboring nodes to the center.
% With the key value in hyperbolic manifold, 
% the cross-geometry attention here is formulated as follows,

\subsubsection*{\textbf{Comparison to Graph Transformers.}}
% We compare with graph convolutional nets and graph transformers.
% %Both of them are typically trained on a specific graph.
% Note that, the parameters of graph convolutional nets are dependent on the specific graph size and feature dimension, preventing the transferability to other graphs.
Despite the differences in generalization capacity, 
%it is noteworthy to mention that 
the proposed architecture is fundamentally different from that of graph transformers,
which conducts the bidirectional attention to all nodes, typically in Euclidean space.
On the contrary, the proposed attention is unidirectional and is performed over graph substructure in account of its Riemannian geometry.
We notice that \citet{kdd24Hypformer} introduces a transformer net (Hypformer) very recently.
However, we consider each substructure in the corresponding Riemannian manifold with cross-geometry keys, while Hypformer places the input as a whole in hyperbolic space.

% \begin{figure}
% \centering
%     \includegraphics[width=0.9\linewidth]{cross-att}
%     \vspace{-0.1in}
%         \caption{Cross-geometry attention in hyperbolic space.}
%     \label{fig:CrossAttention}
%         \vspace{-0.1in}
% \end{figure}

\subsubsection{\textbf{Global Learning Module}}
%It is aimed to induce the node encodings of the graph.
%We aim to learn node encodings of the input graph in this module.
Sampling multiple substructures from the graph, this module examines the entire graph to learn node encodings from a global perspective.
This objective is achieved by the following two phases.

\begin{figure}
    \centering
        \includegraphics[width=0.77\linewidth]{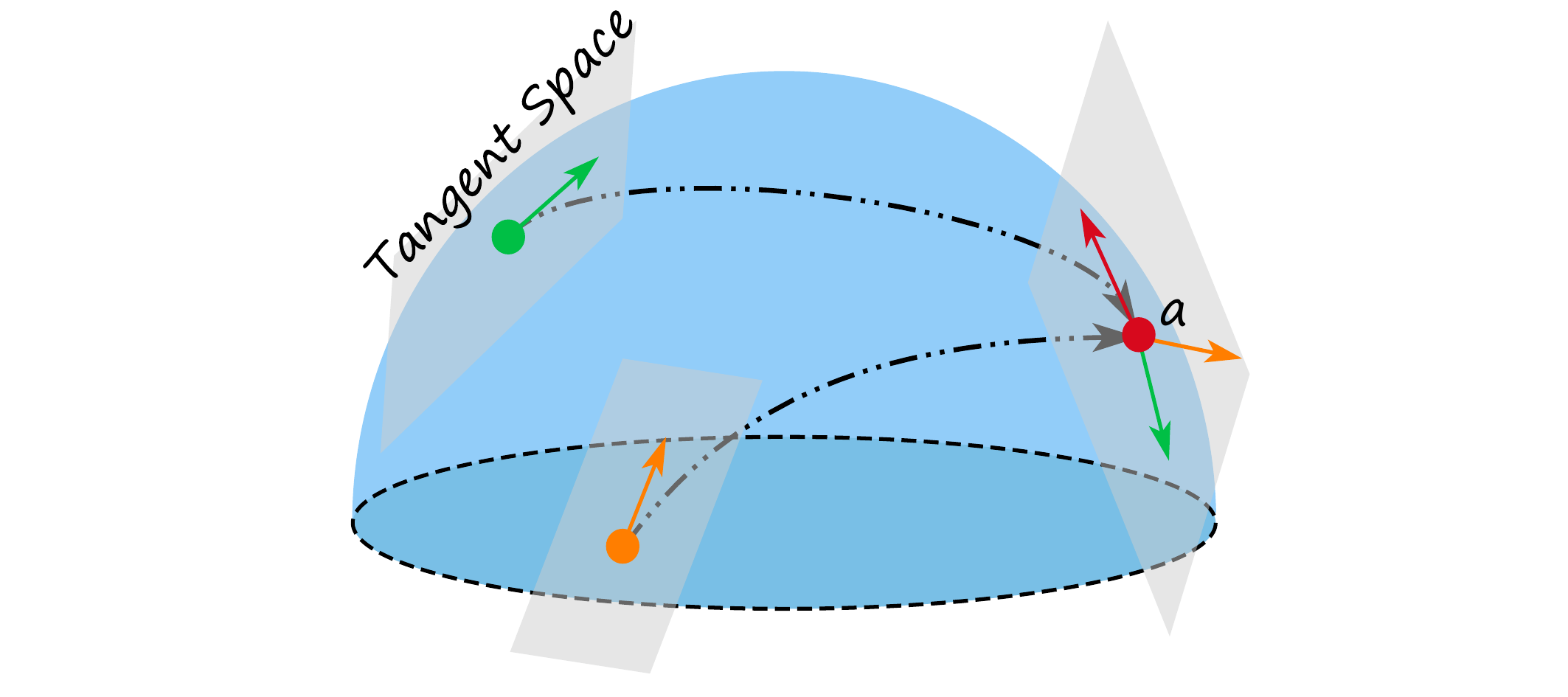}
                \vspace{-0.18in}
            \caption{An illustration of bundle convolution.}
        \label{fig:bundle}
            \vspace{-0.2in}
    \end{figure}

Firstly, we study the node encoding at substructure level.
Note that, 
node encodings live in the tangent bundle surrounding the manifold, where the tangent space of one point is incompatible with that of another point \cite{Petersen16}.
Hence, existing message passing formulations (e.g., GCN \cite{iclr17gcn}, Constant Curvature GCN \cite{icml20Constant}) cannot be used due to space incompatibility.
To bridge this gap, we propose a \textbf{Bundle Convolution} for  message passing over tangent bundles.
The unified formalism for arbitrary curvature is derived as,
\begin{align}
BC_{\boldsymbol p_t}(\{\boldsymbol p_i, \boldsymbol z_i\}_{i\in\Lambda})=\sum\limits_{i\in\Lambda}\left(\alpha_{it} \boldsymbol z_i-\frac{\kappa\alpha_{it}\langle\boldsymbol z_i, \boldsymbol p_t\rangle_\kappa}{1+\kappa\langle\boldsymbol p_i, \boldsymbol p_t\rangle_\kappa}(\boldsymbol p_i+\boldsymbol p_t)\right),
\label{eq:bconv}
\end{align}
% where $\beta_{it}=\frac{\kappa\langle\boldsymbol z_i, \boldsymbol p_t\rangle_\kappa}{1+\kappa\langle\boldsymbol p_i, \boldsymbol p_t\rangle_\kappa}$, and $\boldsymbol p_t$ is the coordinate of the target point. $\boldsymbol M$ is the parameter.
where $\Lambda$ is the node set of the substructure and the attentional weight $\alpha_{it}$ is derived by Eq. (\ref{eq:att}) over the substructure.
The rationale of resolving space incompatibility lies in parallel transport, a canonical way to connect different tangent spaces.
\begin{mymath}
\newtheorem*{def2}{Parallel Transport} 
\begin{def2}
In Riemannian geometry, the parallel transport w.r.t. the Levi-Civita connection $PT_{x\to y}$ transports a vector in $\boldsymbol v \in \mathcal T_x\mathcal M$ to another tangent space $\mathcal T_y\mathcal M$ with a linear isometry along the geodesic between $\boldsymbol x, \boldsymbol y \in \mathcal M$.
\end{def2}
\end{mymath}
\vspace{-0.05in}
\noindent Accordingly,  Eq. (\ref{eq:bconv}) can be explained as the following process. 
The encodings are parallel transported to the tangent space of the target point, in which message passing is subsequently conducted. 
A visual illustration is given in Fig. \ref{fig:bundle}, where $a$ is the target point whose encoding is to be updated. 
The advantage of bundle convolution is that we consider the encoding of the global structure while encapsulating the local geometry of the manifold.
(The detailed derivation is provided  in Appendix C.)

% \begin{figure}
% \centering
%     \vspace{-0.1in}
%     \includegraphics[width=1\linewidth]{bundle}
%         \caption{An illustration of bundle convolution.}
%     \label{fig:bundle}
%         \vspace{-0.1in}
% \end{figure}

Secondly, we obtain the output node encoding at the graph level.
As in \textbf{Fig. \ref{fig:overall}(c)}, there are three cycles (and trees) containing node $a$, and we aim to align the coordinates of node $a$ and obtain the graph-level node encoding.
%Thus,  node coordinate is calculated by  ....
With $K$ samples of $a$, the alignment is given by the geometric midpoint of coordinates on the manifold, $\boldsymbol p_{a}=mid_\kappa(\{\boldsymbol p_{a_i}\}_{i=1,\cdots, K}) \in \mathcal L^d_\kappa$, where aggregation weight is set as $1$.
%Given the aligned coordinate as reference, 
Then, 
node encodings of each sample are parallel transported to the tangent space of the midpoint.
Consequently, the graph-level node encoding is derived as $\boldsymbol z_{a}=BC_{\boldsymbol p_{a}}(\{\boldsymbol p_{a_i},\boldsymbol z_{a_i}\}_{i=1, \cdots, K})$.
% To be specific, , we first make the agreement of the node coordinates, which is given by the geometric midpoint on the manifold. It takes the form as follows,

% Second, we design a hierarchical aggregation mechanism to learn the node encodings in accordance to their coordinates on the manifold.

% We propose to bridge this gap by the parallel transport between different tangent spaces along the geodesic.

% \noindent In hierarchical aggregation mechanism, substructure-level aggregation is followed up with a graph-level one. 
% In each substructure, 
% %we consider the encodings of the ancestor node and the descendant nodes in a tree or the neighboring nodes in a cycle.
% each node is updated in account of its ancestor node and descendant nodes in a tree or the neighboring nodes in a cycle, and node encodings are parallel transported to the tangent space of the node to be updated.
% Subsequently, the graph-level aggregation summarizes the encoding of same node from its samples in different substructures, 
% and 
% The aggregation over the tangent bundle takes the form as follows,

%The objective of this module is to update node coordinates in the substructure () on the factor manifold

        \vspace{-0.05in}
\subsubsection*{\textbf{On Stacking Multiple Layers.}}
%Stacking the proposed Riemannian layer 
The main advantage is to enlarge the receptive field.
For example, a node in the tree is updated by its first-order descendant nodes with one layer, as in Eq. (\ref{eq:cross-geo}),
and is further affected by the second-order descendant nodes with another layer.
%, given the unidirectional attention in Eq. (), 
%The node is further affected by the second-order descendant nodes via stacking another layer.
%The expansion pattern of the receptive field is given by the structural geometry, and is different from that of traditional convolutional layer.
By stacking multiple layers, a node can perceive a larger region in the substructure, 
%while broadening the global view simultaneously by calculating the agreement over the whole graph.
while simultaneously broadening its global view by calculating the agreement across the entire graph.

        \vspace{-0.05in}
\subsubsection*{\textbf{Comparison to Previous Riemannian Graph Models.}}
Our idea is fundamentally different from that of previous Riemannian models.
%The essential difference lies in the output space 
All of them explore advanced techniques to embed nodes in the \textbf{manifold}, either in a single CCS \cite{icml20Constant}, the product \cite{iclr19mixCurvature,iclr23lantentGraphProduct}, or the quotient \cite{nips21UltraNN,nips22QGCN},
%, to the best of our knowledge. 
but we seek node encodings in the \textbf{tangent bundle}, considering the structural vocabulary and global information.

\begin{algorithm}[t]
        \caption{Training Algorithm of  \texttt{RiemannGFM}} 
        \KwIn{pre-training graphs, Hyperparameters of the  product bundle and \texttt{RiemannGFM}.}
        \KwOut{Model parameters of \texttt{RiemannGFM}}
        Initialize node encodings and node coordinates on CCSs;

        Sample substructures according to the structural vocabulary;

\While{model not converged}{   
            \For{each substructure in each geometry} {
               Conduct the cross-geometry attention in Eq. (\ref{eq:cross-geo}) to update node coordinates;

                Conduct the bundle convolution  in Eq. (\ref{eq:bconv}) to update node encodings in the substructure;
                }

                Induce the node encodings with global view with the geometric midpoint for each geometry;

                Generate the hyperbolic and hyperspherical  views;
            
            Compute the geometric contrastive loss with Eq. (\ref{eq:loss});

            Update model parameters via gradient descent.
}
\end{algorithm}

\begin{table*}[ht]
    \caption{Cross-domain transfer learning performance on Citeseer, Pubmed, GitHub and Airport datasets. Node classification and link prediction results are reported. The best results are in \textbf{boldfaced}.}
    \vspace{-0.1in}
    \label{tab:finetune}
    \centering
    \resizebox{1.025\linewidth}{!}{
    \begin{tabular}{c  c | cc cccc c c | cc cccccc}
    \hline
        & & \multicolumn{8}{c|}{\textbf{Node Classification Results}}    & \multicolumn{8}{c}{\textbf{Link Prediction Results}}\\ 
    \cline{3-18}
    & \textbf{Method} & \multicolumn{2}{c}{\textbf{Citeseer}} &  \multicolumn{2}{c}{\textbf{Pubmed}}  &  \multicolumn{2}{c}{\textbf{GitHub}} &  \multicolumn{2}{c|}{\textbf{Airport}}  &  \multicolumn{2}{c}{\textbf{Citeseer}} &  \multicolumn{2}{c}{\textbf{Pubmed}}  &  \multicolumn{2}{c}{\textbf{GitHub}} &  \multicolumn{2}{c}{\textbf{Airport}} \\
    & &ACC & F1 & ACC & F1 & ACC & F1 & ACC & F1  & AUC & AP & AUC & AP & AUC & AP & AUC & AP\\
    \hline
    & GCN \cite{nips19hgcn}
    & 70.30 & 68.56 & 78.90 & 77.83 & 85.68 & 84.34 & 50.80 & 48.09
    & 90.70 & 92.91 & 91.16 & 89.96 & 87.48 & 85.34 & 92.37 & 94.24 \\
    & SAGE \cite{nips17GraphSAGE}
    & 68.24 & 67.60 & 77.57 & 73.61 & 85.12 & 77.36 & 49.16 & 47.57
    & 87.29 & 89.03 & 87.02 & 86.85 & 79.13 & 81.21 & 92.17 & 93.56 \\
    & DGI \cite{iclr19dgi}
    & 71.30 & 71.02 & 76.60 & 76.52 & 85.19 & 84.10 & 50.10 & 49.56
    & 96.90 & 97.05 & 88.39 & 87.37 & 86.39 & 86.61 & 92.50 & 91.63 \\
    & GraphMAE2 \cite{www23graphmae2}
    & \textbf{73.40} & 71.68 & \textbf{81.10} & \textbf{79.78} & 85.23 & 83.34 & 52.34 & 49.02
    & 92.75 & 89.23 & 89.46 & 85.37 & 87.11 & 86.23 & 88.23 & 90.23 \\
    \hline
    \multirow{6}{*}{\rotatebox{90}{GFM}}
    & GCOPE \cite{kdd24gcope}
    & 65.33 & 62.34 & 74.15 & 74.33 & 82.29 & 72.89 & 39.96 & 36.40
    & 88.60 & 83.03 & 90.84 & 86.45 & 82.16 & 83.22 & 86.17 & 84.91\\
    & OFA \cite{iclr24ofa}
    & 58.32 & 65.41 & 74.40 & 72.42 & - & - & - & -
    & 82.62 & 83.74 & 92.26 & 91.36 & - & - & - & - \\
    & GraphAny \cite{zhao2024graphany}
    & 66.10 & 63.01 & 76.10 & 70.12 & 79.45 & 77.19 & 47.98 & 46.88
    & - & - & - & - & - & - & - & - \\
    & OpenGraph\cite{xia2024opengraph}
    & 58.58 & \textbf{76.78 }& 58.40 & 56.49 & 30.16 & 30.16 & 40.45 & 38.28
    & 76.78 & 77.35 & 70.02 & 72.23 & 86.72 & 87.42 & 85.32 & 83.25 \\
    & LLaGA \cite{icml24llaga}
    & 59.00 & 56.91 & 71.21 & 63.38 & 53.33 & 54.17 & 38.49 & 39.89
    & 86.26 & 83.35 & 84.04 & 76.48 & 71.25 & 70.63 & 77.90 & 74.30 \\
    \hline
    & \textbf{\texttt{RiemannGFM}}
    & 66.38  & 66.41  & 76.20  & 75.83  & \textbf{85.96}  & \textbf{85.57}  & \textbf{55.29}  & \textbf{53.27} 
    & \textbf{99.40}  & \textbf{98.42}  & \textbf{94.12} & \textbf{91.64}  & \textbf{89.18} & \textbf{93.52}& \textbf{93.68} & \textbf{96.07}  \\
    % \hline
    % & Improvement
    % & & & & & & & &
    % & & & & & & & & \\
    \hline
    \end{tabular}
    }
        \vspace{-0.1in}
    \end{table*}

\vspace{-0.05in}
\subsection{Geometric Contrastive Learning} \label{sec:gcl}
A foundation model requires self-supervised learning to 
%learn the shared knowledge not specified to the annotations.
acquire shared knowledge that is not tied to specific annotations.
Contrastive learning has become an effective method for self-supervised learning, but it is nontrivial for graphs; 
%on the graphs, e.g., graph augmentation to generate contrastive views is not as easily accessible as the cropping and rotating of images.
for example, graph augmentation to generate contrastive views is not as easily accessible as cropping/rotating of images.
Thanks to the diverse structural geometries in our design,  they offer different views for graph contrastive learning (i.e., hyperbolic view and hyperspherical view).

Here, we introduce the geometric contrastive objective on the product bundle for the self-supervised learning of our model, free of graph augmentation.
Concretely, the node encoding in the tangent space acts as the geometric view of the corresponding manifold.
Thus, the remaining ingredient is a score function that contrasts positive and negative samples, and the challenge lies in the incompatibility between different geometries.
To bridge this gap, we consider a shared tangent space of the north pole of Lorentz/Spherical model.
Thus, the geometric contrast is given as follows,
  \vspace{-0.07in}
\begin{align}
\label{eq:loss}
\resizebox{0.922\hsize}{!}{$
    \mathcal{J}(H, S) = -\sum_{i=1}^N \log\frac{\exp(\langle PT_{\boldsymbol p^H_i \to \boldsymbol o}(\boldsymbol z^H_i), PT_{\boldsymbol p^S_i \to \boldsymbol o}(\boldsymbol z^S_i)\rangle)}{\sum_{j=1}^N \exp(\langle PT_{\boldsymbol p^H_i \to \boldsymbol o}(\boldsymbol z^H_i), PT_{\boldsymbol p^S_j \to \boldsymbol o}(\boldsymbol z^S_j)\rangle)}.
    $}
      \vspace{-0.12in}
\end{align}
The overall objective is formulated as  $\mathcal{J}_0= \mathcal{J}(H, S)+\mathcal{J}(S,H)$, where $N$ is the number of nodes.
Though the geometric contrast is done over node encodings in tangent spaces, the parameters of factor manifolds are encapsulated in the parallel transport among tangent spaces.
The training procedure is summarized in Algorithm 1, 
whose \textbf{computational complexity} is yielded as $O(|\mathcal V|^2+|\mathcal E|)$, where $\mathcal V$ and  $\mathcal E$ are the node set and edge set, respectively, and the proposed model supports minibatch training.
Finally, \textbf{\texttt{RiemannGFM}
is capable of generating informative node encodings for an arbitrary graph, with the shared structural knowledge of }\textbf{the graph domain learned in Riemannian geometry.}

\vspace{-0.1in}
\section{Experiment}
  \vspace{-0.03in}
We aim to answer the following research questions:
% The experiments are aimed at answering the following research questions:
\emph{RQ1.} How does \texttt{RiemannGFM} perform in  cross-domain  transfer learning?
\emph{RQ2.} How significant is embedding structural vocabulary into Riemannian geometry, rather than Euclidean ones? 
\emph{RQ3.} How effective is  \texttt{RiemannGFM} under few-shot learning?
\emph{RQ4.} How expressive is the structural knowledge learned by \texttt{RiemannGFM}?
\emph{RQ5.} How does the pre-training dataset impact  \texttt{RiemannGFM}?

\vspace{-0.05in}
\subsection{Experimental Setups}

\subsubsection{\textbf{Datasets}}
The experiments are conducted on a diversity of datasets. 
Specifically, we include two text-attributed graphs (the popular Citeseer and Pubmed \cite{citeseerandpubmed}),
one mix-attributed graph (GitHub \cite{github}),
and one non-attributed graph  (Airports \cite{kdd17struc2vec}).
%, are text-attributed graph datasets from different fields, 
% (3) GitHub \cite{github} is a social network \textbf{graph with mixed attributes} consisting of location, starred repositories, employer, and e-mail address. A \textbf{non-attributed graph} of  (4) USA Airports \cite{kdd17struc2vec} contains airports and the connections among them. 
%labels correspond to activity levels.
%To ensure a comprehensive evaluation, we compare our proposed framework against three distinct categories of models. 

\subsubsection{\textbf{Baselines \& Metrics}}
We include $9$ strong baselines categorized into three groups:
The first group is the \textbf{vanilla GNNs}: GCN \cite{iclr17gcn} and GraphSAGE \cite{nips17GraphSAGE} with the end-to-end training paradigm.
% These models leverage neighborhood aggregation techniques to iteratively gather information from adjacent nodes, employing an end-to-end training paradigm without the necessity for pre-training. 
The second group consists of \textbf{self-supervised graph learning models}: DGI \cite{iclr19dgi} and GraphMAE2 \cite{www23graphmae2}. 
%The GSSL methods emphasize the transferability of learned structural representations without relying on explicit label information. 
The third group is \textbf{graph foundation models}, including OFA \cite{iclr24ofa}, GCOPE \cite{kdd24gcope}, GraphAny \cite{zhao2024graphany}, LLaGA \cite{icml24llaga} and OpenGraph \cite{xia2024opengraph}. 
%The proposed model is named \texttt{RiemannGFM}, considering the shared structural knowledge.
%These models could train on extensive and diverse datasets to enhance cross-task and cross-domain generalization.
We evaluated the comparison methods by both node classification and link prediction tasks.
%For our evaluation, we employed two widely adopted performance metrics for each of the node classification and link prediction tasks. 
For node classification, we employ two popular metrics of classification accuracy (ACC) and weighted F1-score (F1), while for the link prediction,  AUC and Average Precision (AP) are utilized. 
% To ensure statistical robustness and fair comparison, we conducted $10$ independent runs for each model and reported the mean performance along with the standard deviation.
% All experiments are conducted on a server equipped with an NVIDIA GeForce RTX 4090D GPU (24GB VRAM) utilizing CUDA 11.8, an AMD EPYC 9754 128-Core Processor (18 vCPUs allocated), 60GB of RAM.
%, running Ubuntu 20.04.5 LTS as the operating system.

\subsubsection{\textbf{Model Configuration and Reproducibility.}}
% In the proposed \texttt{RiemannGFM}, the input node encoding is given from the Laplacian matrix, encapsulating the structural information.
% Concretely, we leverage the eigenvectors of $K$ largest eigenvalues for the normalization among different graph datasets.
% The initial node coordinates are given by the normalized vectors.
% In particular, the normalized vectors are projected to the manifold-valued coordinates 
As for the initialization,
% of the proposed \texttt{RiemannGFM} as follows.
the input node encoding is given from the Laplacian matrix, encapsulating the structural information.
Note that, we leverage the eigenvectors of $K$ largest eigenvalues, which normalizes different graph datasets with a predefined $K$.
Correspondingly, node coordinates are initialized on the constant curvature spaces by the exponential map with the reference point of the north pole.
On model configuration,
we utilize the standard curvature for hyperbolic and hyperspherical spaces, and the dimension is set as $32$ by default. That is, \texttt{RiemannGFM} is instantiated on the product bundle of 
$\left(\mathcal H^{32}_{-1} \otimes \mathcal T\mathcal H^{32}_{-1}\right)$
$ \otimes \left(\mathcal S^{32}_{1} \otimes \mathcal T\mathcal S^{32}_{1}\right)$.
The \texttt{RiemannGFM} consists of two universal Riemannian layers.
As for the structural vocabulary, trees are in hyperbolic space, while hyperspherical space accommodates cycles of node triangles and quadruples.
%The parameters are optimized by Adam \cite{iclr15Adam} with the learning rate and dropout rate are tuned with grid search.
\textbf{Codes} are available at {\color{blue}\url{https://github.com/RiemannGraph/RiemannGFM}}.
%{\url{https://anonymous.4open.science/r/Geo-GFM-1603}}. 
(Implementation notes are in Appendix E.)

\begin{table}[t]
\centering % Center the table within the half-column width
\caption{Geometric ablation on Citeseer, Pubmed, and Airport datasets. Link prediction results are reported in terms of AUC (\%). The results are given in the form of mean$\pm$std. $\mathcal R^{32}_{0}$ denotes the Euclidean space.}
       \vspace{-0.1in}
\label{tab:geo-ablation}
\begin{tabular}{cc|c c c}
\hline
\textbf{Trees} & \textbf{Cycles} & \textbf{Citeseer} & \textbf{Pubmed} & \textbf{Airport} \\
\hline
$\mathcal H^{32}_{-1}$ & $\mathcal S^{32}_{1}$ & \textbf{99.40 $\pm$ 0.06} & \textbf{94.12 $\pm$ 1.38} & \textbf{93.68 $\pm$ 0.09} \\
$\mathcal H^{32}_{-1}$ & $\mathcal R^{32}_{0}$ & 98.48 $\pm$ 0.32 & 92.41 $\pm$ 2.14 & 92.22 $\pm$ 1.43 \\
$\mathcal H^{32}_{-1}$ & $\mathcal H^{32}_{-1}$ & 98.21 $\pm$ 0.45 & 92.32 $\pm$ 2.57 & 91.79 $\pm$ 1.58 \\
\hline
$\mathcal H^{32}_{-1}$ & $\mathcal S^{32}_{1}$ & \textbf{99.40 $\pm$ 0.06} & \textbf{94.12 $\pm$ 1.38} & \textbf{93.68 $\pm$ 0.09} \\
$\mathcal R^{32}_{0}$ & $\mathcal S^{32}_{1}$ & 98.72 $\pm$ 0.08 & 92.43 $\pm$ 1.53 & 92.51 $\pm$ 0.18 \\
$\mathcal S^{32}_{1}$ & $\mathcal S^{32}_{1}$ & 98.85 $\pm$ 0.09 & 92.88 $\pm$ 1.47 & 92.85 $\pm$ 0.23 \\
\hline
\end{tabular}
       \vspace{-0.2in}
\end{table}

\vspace{-0.12in}
\subsection{Results and Discussion}
\vspace{-0.01in}
\subsubsection{\textbf{Cross-domain Transfer Learning Performance (RQ1)}}
The results for both node classification and link prediction are summarized in Table \ref{tab:finetune}. 
The proposed \texttt{RiemannGFM} is pre-trained on the datasets of ogbn-arxiv \cite{nips2020arxiv}, Physics \cite{physics_computers}, Amazon-Computers \cite{physics_computers} and its classification head is the same as that of popular GFMs \cite{xia2024opengraph,xia2024anygraph,kdd24gcope}.
Note that, OFA \cite{iclr24ofa} cannot work on the graphs without textual attributes (i.e., Github and Airport datasets). 
GraphAny \cite{zhao2024graphany} generates classification logistics only, and thereby cannot be utilized for link prediction.
As shown in Table \ref{tab:finetune}, in the link prediction task, the proposed \texttt{RiemannGFM} consistently achieves the best results among GFMs and specialized models, i.e., GCN, SAGE, DGI and GraphMAE2.
The structural vocabulary embedded in Riemannian manifolds contributes to our success, which is further discussed  in RQ3.
In node classification tasks, the proposed \texttt{RiemannGFM} achieves the best results on  the graphs without textual attributes. 
On text-attributed graphs,  \texttt{RiemannGFM} still obtain comparable performance to previous GFMs.
This shows the importance of building GFM that can capture the structural information.

\begin{table*}[ht]
    \caption{Few-shot learning performance on Citeseer, Pubmed, GitHub, and Airport datasets.  The best results are in \textbf{boldfaced}.}
    \vspace{-0.1in}
    \label{tab:fewshot_nc}
    \centering
    % \begin{tabular}{c  c | cc cc cc cc}
    \begin{tabular}{c  c |cc cc cc cc}
    \hline
    & &\multicolumn{8}{c}{\textbf{Node Classification Results}}  \\ 
    \cline{3-10}
    \textbf{Setting} & \textbf{Method} & \multicolumn{2}{c}{\textbf{Citeseer}} &\multicolumn{2}{c}{\textbf{Pubmed}}  & \multicolumn{2}{c}{\textbf{GitHub}} & \multicolumn{2}{c}{\textbf{Airport}}  \\
    % \multicolumn{2}{c|}{\textbf{Method} } & \textbf{Citeseer} & \textbf{Pubmed}  & \textbf{GitHub} & \textbf{Airport}  & \textbf{Citeseer} & \textbf{Pubmed}  & \textbf{GitHub} & \textbf{Airport} \\
    & &ACC & F1 & ACC & F1 & ACC & F1 & ACC & F1  \\
    \hline
    \multirow{7}{*}{1-Shot}
    & DGI \cite{iclr19dgi}
    & 37.40 {\scriptsize$\pm$9.98} & 32.29 {\scriptsize$\pm$12.17} & 39.29 {\scriptsize$\pm$3.79} & 34.76 {\scriptsize$\pm$5.12} & 59.90 {\scriptsize$\pm$4.89} & 55.48 {\scriptsize$\pm$9.73} & 30.63 {\scriptsize$\pm$6.14} & 17.57 {\scriptsize$\pm$7.28} \\
    & GraphMAE2 \cite{www23graphmae2}
    & 34.62 {\scriptsize$\pm$4.23} 
    & 31.34 {\scriptsize$\pm$1.21} 
    & 39.10 {\scriptsize$\pm$6.45} 
    & 35.97 {\scriptsize$\pm$8.83}
    & 52.47 {\scriptsize$\pm$3.98} 
    & 50.25 {\scriptsize$\pm$4.78} 
    & 29.89 {\scriptsize$\pm$5.45} 
    & 20.27 {\scriptsize$\pm$6.51} \\
    & OFA \cite{iclr24ofa}
    & 37.58 {\scriptsize$\pm$10.51} 
    & 30.90 {\scriptsize$\pm$2.85} 
    & 39.80 {\scriptsize$\pm$0.74}
    & 27.54 {\scriptsize$\pm$3.05}
    & -
    &- 
    &- 
    &- \\
    & GCOPE \cite{kdd24gcope}
    & 36.03 {\scriptsize$\pm$4.63} 
    & 31.89 {\scriptsize$\pm$4.54} 
    & 37.36 {\scriptsize$\pm$4.21} 
    & 23.64 {\scriptsize$\pm$3.80} 
    & 56.07 {\scriptsize$\pm$5.09} 
    & 43.89 {\scriptsize$\pm$6.22} 
    & 26.09 {\scriptsize$\pm$0.99} 
    & 18.05 {\scriptsize$\pm$4.95} \\
    & OpenGraph \cite{zhao2024graphany}
    & 20.60 {\scriptsize$\pm$2.43} & 18.30 {\scriptsize$\pm$1.01} & 43.58 {\scriptsize$\pm$1.12} & 35.39 {\scriptsize$\pm$1.03} & 22.19 {\scriptsize$\pm$0.03} & 40.32 {\scriptsize$\pm$0.65} & 31.94 {\scriptsize$\pm$2.99} & 23.38 {\scriptsize$\pm$2.13} \\
    & LLaGA \cite{icml24llaga}
    & 18.10 {\scriptsize$\pm$2.03} & 14.57 {\scriptsize$\pm$0.97} & 35.68 {\scriptsize$\pm$1.58} & 33.48 {\scriptsize$\pm$1.34} & 26.67 {\scriptsize$\pm$1.96} & 28.89 {\scriptsize$\pm$2.54} & 23.53 {\scriptsize$\pm$2.02} & 19.17 {\scriptsize$\pm$2.31} \\
    & \textbf{\texttt{RiemannGFM}}  (Ours)
    & \textbf{38.02 {\scriptsize$\pm$9.45}} & \textbf{32.42 {\scriptsize$\pm$9.87}} & \textbf{45.24 {\scriptsize$\pm$3.55} }& 
    \textbf{37.87} {\scriptsize$\pm$6.56} & \textbf{77.83 {\scriptsize$\pm$4.53}} & \textbf{72.46 {\scriptsize$\pm$7.54}} & \textbf{32.61 {\scriptsize$\pm$4.74}} & \textbf{27.18 {\scriptsize$\pm$7.46}} \\
    \hline
    \multirow{7}{*}{5-Shot}
    & DGI \cite{iclr19dgi}
    & 46.48 {\scriptsize$\pm$1.32} & 43.62 {\scriptsize$\pm$1.49} & 51.38 {\scriptsize$\pm$4.05} & 50.90 {\scriptsize$\pm$3.86} & 65.38 {\scriptsize$\pm$0.13} & 64.55 {\scriptsize$\pm$0.28} & 37.61{\scriptsize$\pm$6.41} & 28.85 {\scriptsize$\pm$6.16} \\
    & GraphMAE2 \cite{www23graphmae2}
    & 47.12 {\scriptsize$\pm$4.01} & 44.71 {\scriptsize$\pm$1.88} & 53.04 {\scriptsize$\pm$4.11} & 47.74 {\scriptsize$\pm$4.37} & 62.22 {\scriptsize$\pm$2.19} & 60.88 {\scriptsize$\pm$7.42} & 37.09 {\scriptsize$\pm$6.02} & 29.11 {\scriptsize$\pm$2.02} \\
    & OFA \cite{iclr24ofa}
    & 31.90 {\scriptsize$\pm$4.27} & 23.04 {\scriptsize$\pm$0.83} & 36.72 {\scriptsize$\pm$9.40} & 24.43 {\scriptsize$\pm$6.12} 
    &- &- & -& -\\
    & GCOPE \cite{kdd24gcope}
    & 43.48 {\scriptsize$\pm$9.55} & 38.65 {\scriptsize$\pm$9.07} & 46.35 {\scriptsize$\pm$9.59} & 44.85 {\scriptsize$\pm$9.36} & 73.26 {\scriptsize$\pm$2.07} & 63.13 {\scriptsize$\pm$1.57} & 33.18 {\scriptsize$\pm$2.38} & 27.71 {\scriptsize$\pm$6.09} \\
    & OpenGraph \cite{zhao2024graphany}
    & 29.30 {\scriptsize$\pm$1.81} & 27.46 {\scriptsize$\pm$1.41} & 37.52 {\scriptsize$\pm$2.52} & 35.51 {\scriptsize$\pm$3.24} & 24.32 {\scriptsize$\pm$0.45} & 42.30 {\scriptsize$\pm$0.04} & 33.51 {\scriptsize$\pm$3.55} & 23.74 {\scriptsize$\pm$2.15} \\
    & LLaGA \cite{icml24llaga}
    &21.60 {\scriptsize$\pm$2.11} & 24.89 {\scriptsize$\pm$2.32} &  32.02 {\scriptsize$\pm$1.85} &  {35.84 \scriptsize$\pm$1.96} & 58.33 {\scriptsize$\pm$1.35} & 56.86 {\scriptsize$\pm$1.26} & 32.86 {\scriptsize$\pm$1.24} & 30.50 {\scriptsize$\pm$1.23} \\
    & \textbf{\texttt{RiemannGFM}}  (Ours)
    &\textbf{53.46 {\scriptsize$\pm$4.17}} & \textbf{51.89 {\scriptsize$\pm$4.60}} &  \textbf{66.18 {\scriptsize$\pm$5.99} }&  \textbf{64.56 {\scriptsize$\pm$9.38} }& \textbf{84.19 {\scriptsize$\pm$1.05} }& \textbf{83.13 {\scriptsize$\pm$1.89} }& \textbf{38.72 {\scriptsize$\pm$5.98}} & \textbf{33.40 {\scriptsize$\pm$5.66} }\\
    \hline
    \end{tabular}
    \end{table*}

\begin{figure}[t]
    \centering
    \begin{subfigure}{0.49\linewidth}
        \centering
        \includegraphics[width=\linewidth]{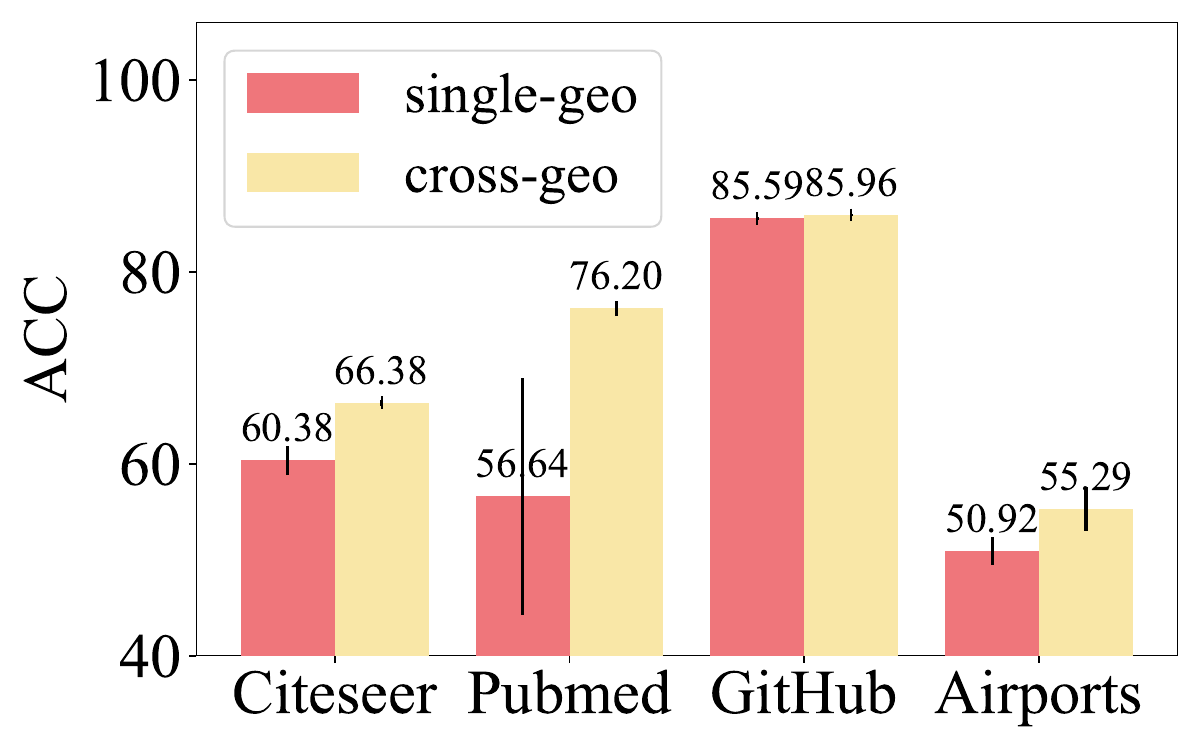}
        \caption{Node Classification}
        \label{fig:chart3}
    \end{subfigure}
    \begin{subfigure}{0.49\linewidth}
        \centering
        \includegraphics[width=\linewidth]{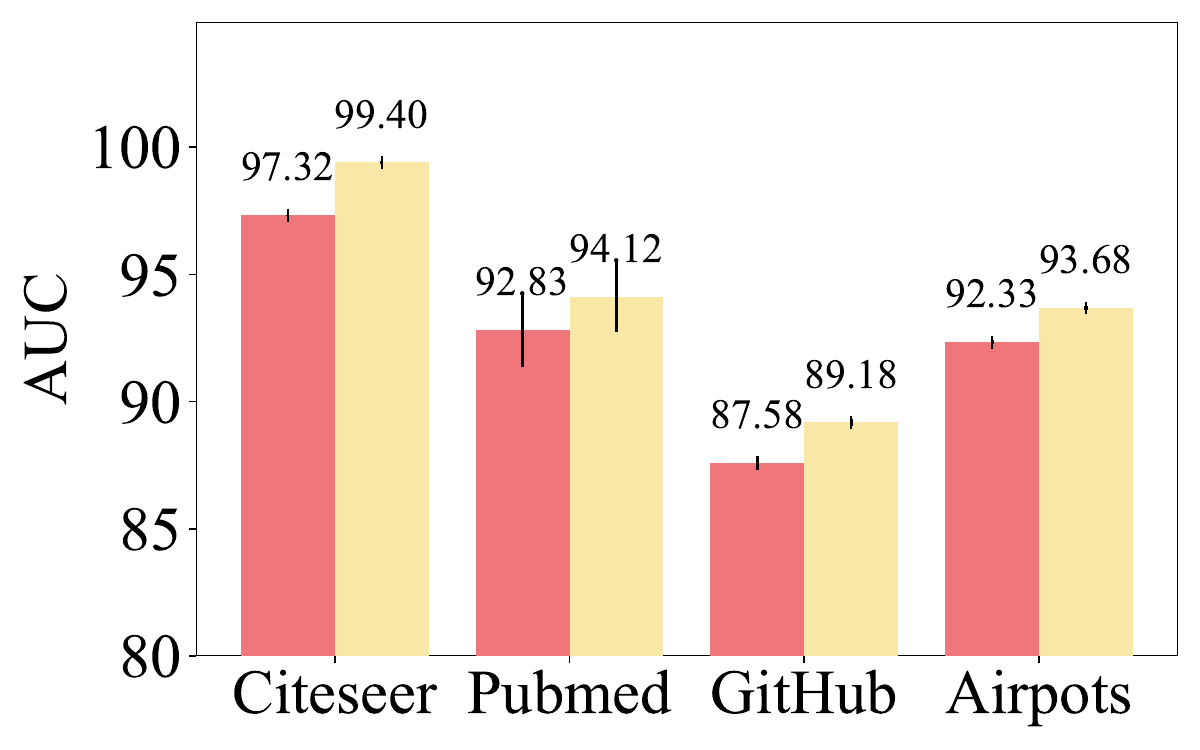}
        \caption{Link Prediction}
        \label{fig:chart2}
    \end{subfigure}
        \vspace{-0.15in}
    \caption{Ablation on cross-geometric attention.}
    \label{fig:attention}
    \vspace{-0.15in}
\end{figure}

\vspace{-0.07in}
\subsubsection{\textbf{The Connection between Structural Vocabulary and Constant Curvature Spaces (CCSs) (RQ2)}}
Given the significance of structural vocabulary, we are interested in which CCS is suitable to model trees and cycles.
Theoretically, 
hyperbolic space aligns with the tree as evidenced in the consistency of volume growth \cite{icml18HierarchiesofHyperbolic},
while hyperspherical space acts as the geometric analogy of cycles according to the common rotational invariant \cite{Petersen16}.
Here, we empirically investigate our choice by geometric ablation.
To be specific, we place trees and cycles in hyperbolic, hyperspherical, and Euclidean spaces, respectively,
and summarize and link prediction results in Table \ref{tab:geo-ablation}.
It achieves the best performance when trees are embedded in hyperbolic space, and cycles in hyperspherical space, aligning with our choices.

%The results in Table X empirically explain our choice. 

% In \texttt{RiemannGFM}, we embed basic hierarchical and cyclical patterns into hyperbolic space and hyper-spherical space respectively. But, how does it perform if basic patterns are embedded into Euclidean space? The results are given in Table \ref{tab:geo-ablation}. We find that the Riemannian Embedding is much superior to the Euclidean one.
% The main reason is that the Riemannian manifolds are endowed with the inherent structural connotation, e.g., hierarchy is inherent in hyperbolic space. Even if two different structures share the same pattern, their geometric meanings differ a lot. For instance, two tree structures with different root nodes will be embedded into different neighborhoods on the manifold, and their difference not only reflect on different locations but also the tangent space of the points. When it comes to Euclidean space, their difference in tangent space part vanishes \emph{since all tangent spaces on points in Euclidean space are isomorphic to the whole space}.
% With intrinsic geometric properties, our \texttt{RiemannGFM} can easily recognize different structural patterns in a more meaningful aspect, and generalize the structural knowledge to node classification and link prediction task.

       \vspace{-0.05in}
\subsubsection{\textbf{Ablation Study on Cross-geometry Attention}}
We conduct an ablation study to evaluate the effectiveness of cross-geometry attention, 
whose query vector is in the counterpart CCS of key and value vector.
To this end, we introduce a model of a single-geometry variant, which utilizes the key, query, and value vectors in the same CCS.
Fig. \ref{fig:attention} collects node classification and link prediction results on different datasets.
The cross-geometry attention consistently outperforms the single-geometry variant, demonstrating the effectiveness of our design.

       \vspace{-0.05in}
\subsubsection{\textbf{Few-shot Learning Performance (RQ3)}}
% We compare our proposed method \texttt{RiemannGFM} with baselines on node classification and link prediction tasks under 1-shot and 5-shot learning. We repeat the evaluation 5 times and report the average results and standard deviation in Table \ref{tab:fewshot}. A detailed explanation of the few-shot method is in Appendix X.
% Non-pre-training methods like GNNs are trained merely on the limited few-shot training data, while other baseline models that use a pre-train-and-tuning paradigm first undergo pre-training before being fine-tuned on the few-shot set. In the pre-training period, we use the pre-rained model they provide initially. During the subsequent tuning phase, any parameters that cannot be directly transferred between datasets are re-learned. 
% We make the following observations:
% Self-supervised methods perform better than Graph foundation-based methods, a phenomenon known as negative transfer. This phenomenon is attributed to the apparent differences in data distribution across cross-domain datasets.
% Our proposed \texttt{RiemannGFM} methods significantly outperform almost all baselines, especially on non-attributed datasets. Because our model can learn structure rather than utilize the structure. When attribute information is reduced or even disappears, our method shows excellent generalization ability.
% Since our \texttt{RiemannGFM} has drawn much common structural knowledge from pre-training datasets, it will learn new structural knowledge fast and distinguish nodes from the structural pattern when meeting a new structure.

We report the node classification results under 1-shot and 5-shot learning in Table \ref{tab:fewshot_nc}. 
The self-supervised models (i.e., DGI and GraphMAE2) are trained merely on the few-shot set, 
while the GFMs undergo model pre-training and are subsequently fine-tuned on the few-shot, following the setting of \cite{xia2024opengraph}. (Further details are introduced in Appendix E.)
As shown in Table \ref{tab:fewshot_nc}, we observe an interesting phenomenon:
OpenGraph and LLaGA exhibit negative transfer on GitHub and Airport datasets.
They leverage the LLM and enjoy shared knowledge among textural attributes. 
However, it becomes problematic when transferring such knowledge to mixed attributes (e.g., the numbers and addresses in GitHub) or to the graphs  without attributes.
This highlights the limitation of coupling graph transfer with textual attributes, and thus supports our motivation  to explore common structures for better universality.
%On the contrary, we highlight the

\begin{figure}[t]
    \centering 
    \begin{subfigure}{0.3\linewidth}
        \centering
        \includegraphics[width=\linewidth]{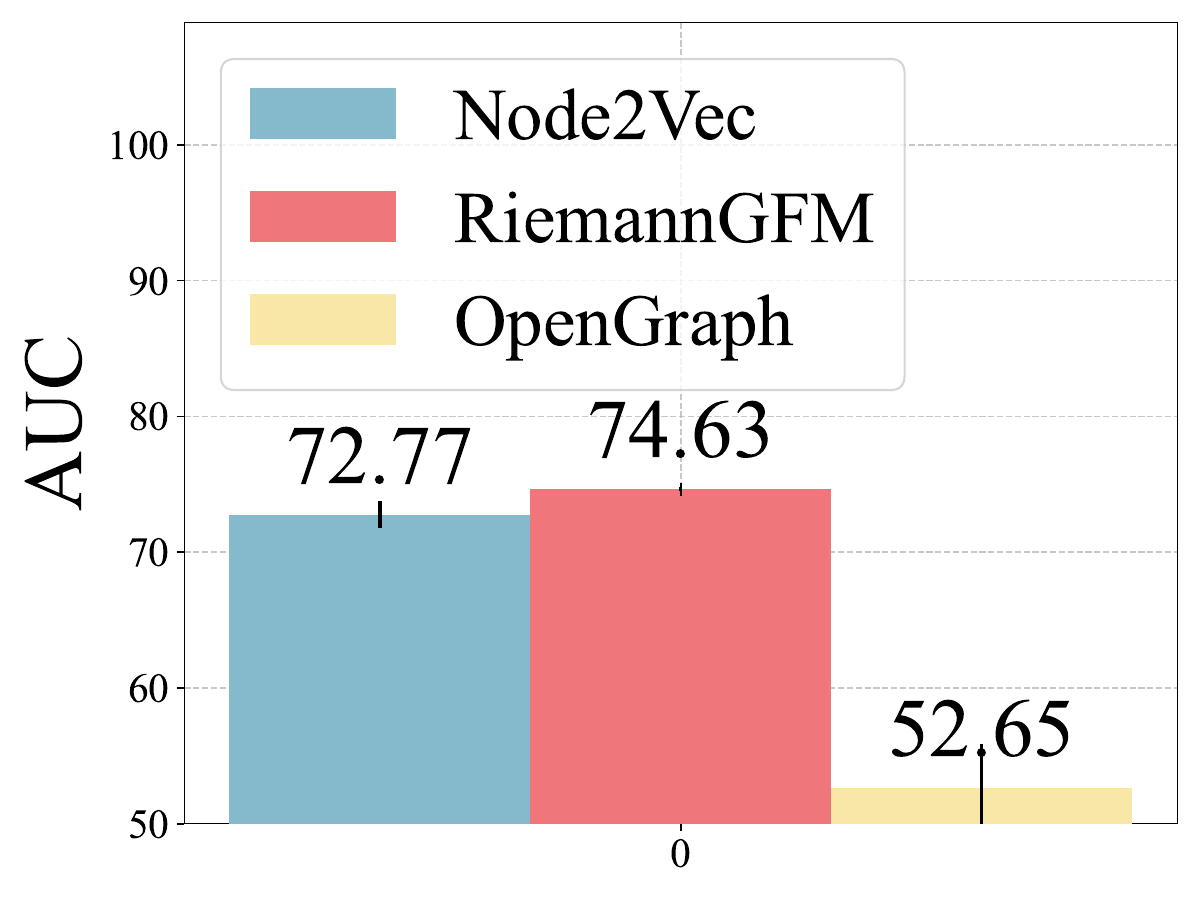}
        \caption{Airports}
        \label{fig:chart1}
    \end{subfigure}
    \begin{subfigure}{0.3\linewidth}
        \centering
        \includegraphics[width=\linewidth]{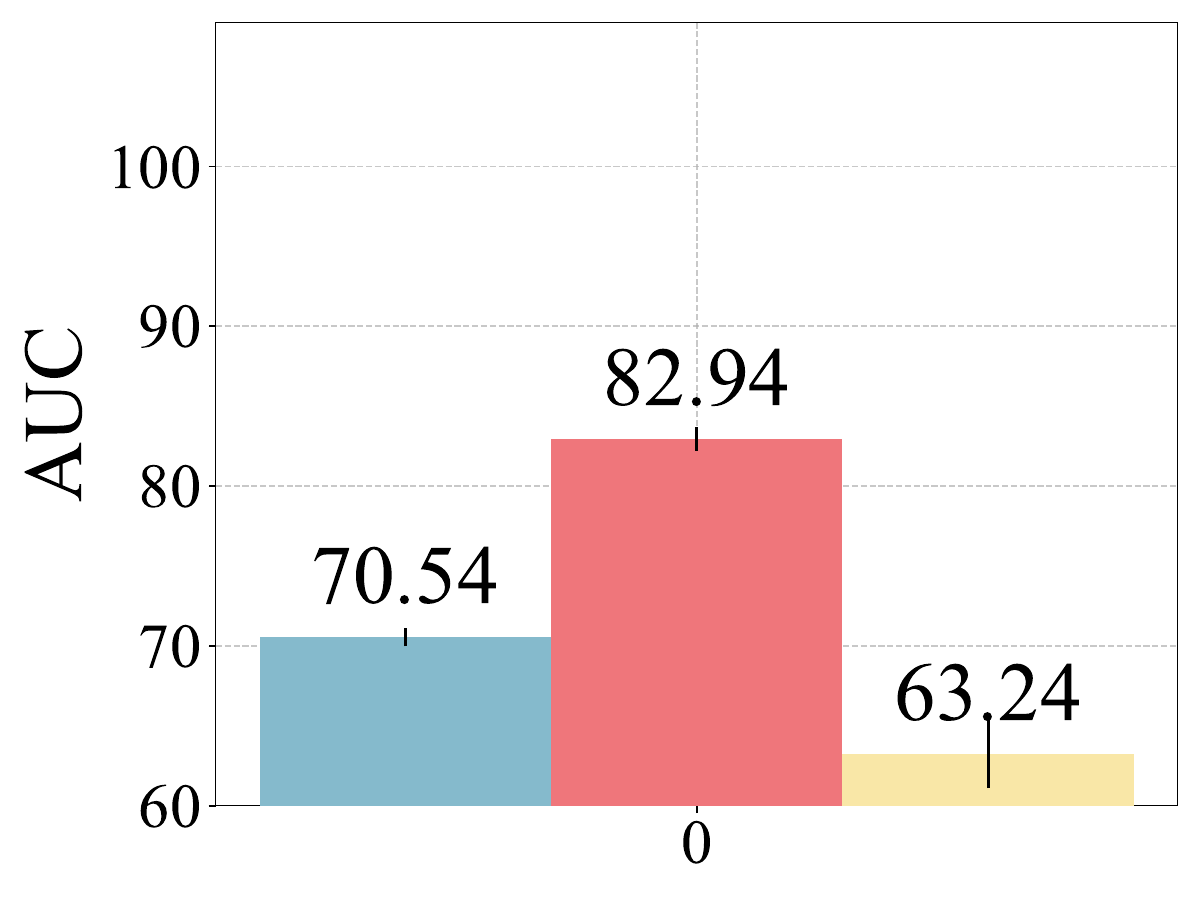}
        \caption{Pubmed}
        \label{fig:chart2}
    \end{subfigure}
    \begin{subfigure}{0.3\linewidth}
        \centering
        \includegraphics[width=\linewidth]{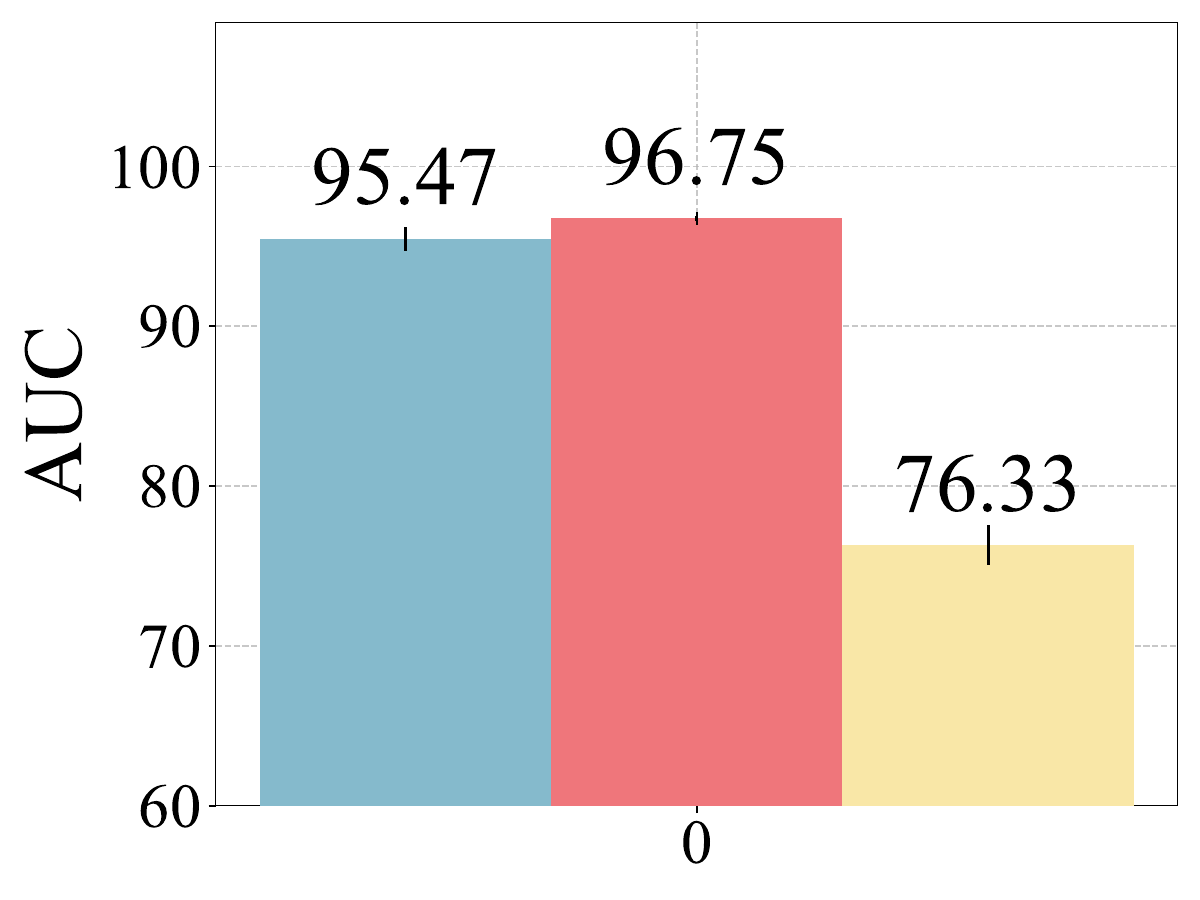}
        \caption{Citeseer}
        \label{fig:chart3}
    \end{subfigure}
           \vspace{-0.15in}
    \caption{Link prediction results with structural knowledge}
    \label{fig:structure-only}
           \vspace{-0.15in}
\end{figure}
  \vspace{-0.05in}
\subsubsection{\textbf{On the Expressiveness of Structural Knowledge (RQ4)}}
We show that, despite the universality of structures in the graph domain, the structural knowledge itself presents promising expressiveness. 
Concretely, we examine the link prediction performance of \texttt{RiemannGFM}, compared to node2vec \cite{kdd16node2vec} 
and GFMs, i.e., OpenGraph \cite{xia2024opengraph}.
In this case, node encodings generated from pre-trained \texttt{RiemannGFM} are utilized for link prediction; that is, we leverage the structural knowledge learned on pre-training datasets and do not include attributes of the target graph, while node2vec is trained on the target datasets.
The results are given in Fig. \ref{fig:structure-only}.
Note that, structural knowledge of \texttt{RiemannGFM} acquires competitive even superior results to specialized model for specific graphs and GFMs incorporated with attributes, showing its promising expressiveness.

% The topology of graph structures is crucial in graph data mining, particularly when nodes lack attributes so that only the structure itself reveals the underlying information. Various methods, such as Spectral Clustering \cite{icml20SpectralClustering}, DeepWalk \cite{kdd14deepwalk}, and Node2Vec \cite{kdd16node2vec}, leverage these structural patterns to extract insights. A key feature of these approaches is their reliance solely on the graph's topology, without node attributes. 
% The proposed \texttt{RiemannGFM} still follows these ideas and can generalize structures to unseen graphs. In Table \ref{tab:finetune}, we assess the performance on non-attributed graphs. Furthermore, we apply structural knowledge exclusively to tasks such as node classification and link prediction in attributed graphs. The results presented in Table X indicate that the expressiveness derived solely from structural features is on par with that obtained from attributed features.
% The graph structure is the complex combination of two basic patterns: hierarchical pattern and cyclical pattern, our proposed \texttt{RiemannGFM} can capture these basic patterns and learn the ability to analyze them automatically when meeting unseen graphs. However, if the hierarchical pattern and cyclical patterns are learned separately, it is still tough to capture the complex structural patterns. Benefiting from $\operatorname{CrossGeo-Attention}$, \texttt{RiemannGFM} can update geometric location interactively between the two basic patterns, considering them together to obtain richer structural information.

\subsubsection{\textbf{Impact of Pre-training Datasets (RQ5)}}
To further investigate the transferability, we study the performance of \texttt{RiemannGFM} with different pre-training datasets.
We adopt Flicker \cite{flickr}, Amazon-Computers \cite{physics_computers}, and WikiCS \cite{wikics} as pre-training datasets respectively, and report the results in Table \ref{tab:diff-train-lp}. We find that: 
\texttt{RiemannGFM} shows more stable performance over different pre-training graphs, that is, the pre-training datasets have limited impact on our model.
However, GCOPE and OpenGraph have higher requirements for pre-training datasets.
Pre-training on similar domains enhances the performance of downstream tasks. For example, when tested on the citation network of Citeseer, OpenGraph achieves 89.64\% with the pre-training dataset WikiCS (citation network),  but has performance loss with pre-training datasets of other domains (65.16\% on Flickr and 60.16\% on AComp).
GCOPE is potentially affected by differences in attribute distribution across different domains.
The reason is two-fold: 
1) The structural transferability of  \texttt{RiemannGFM} enjoys greater universality, especially when attributes show obvious disparities across in different domains.
2) The structural vocabulary is proposed to learn the shared structural knowledge underlying the graph domain and is not tied to any specific structures.
% To assess the efficacy of the knowledge distillation process within our model, we opted to evaluate its performance on a suite of downstream tasks, after its pre-training phase across diverse datasets. 

% Then, we use the pre-trained models obtained from these three different datasets to perform node classification and link prediction experiments on the citation network and social network datasets, respectively. Table \ref{tab:diff-train} shows the impact of datasets from different domains in the pre-training phase on downstream performance. We can observe that pre-training on a dataset from the same domain enhances the performance of downstream tasks for LLaGA and GCN. However, our proposed model does not have special requirements for pre-trained datasets. This shows that the text-attributed graph foundation model has higher requirements for the pre-trained dataset domain and they still need to adjust downstream tasks for cross-domain transfers.

% This advantage is thanks to the structural-transfer ability of \texttt{RiemannGFM}. 

% Even when graphs hail from disparate realms, the vast chasm in their attributes fails to sway the grasp of structural insight. This is because \texttt{RiemannGFM} is designed to assimilate the analytical approach to structures, rather than memorizing the structures themselves.

\begin{table}[t]
    \caption{Cross-domain link prediction prformance on different pre-training datasets.}
    \vspace{-0.1in}
    \label{tab:diff-train-lp}
    \centering
    \begin{tabular}{c  c | cc  c  }
    \hline
    & & \multicolumn{3}{c}{\textbf{Testing Datasets}}\\
    \textbf{Pre-training} & \textbf{Method}  & \textbf{Citeseer} & \textbf{Pubmed}  & \textbf{Airport}  \\
    \hline
    \multirow{3}{*}{Flickr}
    & OpenGraph 
    & 65.16\scriptsize{$\pm$2.54} 
    & 50.66\scriptsize{$\pm$2.14} 
    & 86.42\scriptsize{$\pm$2.15} \\
    & GCOPE &  84.20\scriptsize{$\pm$0.12} &85.60\scriptsize{$\pm$0.62} & 84.54\scriptsize{$\pm$1.09} \\
    & \textbf{\texttt{RiemannGFM}}  &\textbf{99.33\scriptsize{$\pm$1.36} }&\textbf{92.51\scriptsize{$\pm$0.73} }& \textbf{93.52\scriptsize{$\pm$0.06}} \\
    \hline
    \multirow{3}{*}{AComp}
    & OpenGraph 
    & 60.16\scriptsize{$\pm$3.21} 
    & 60.56\scriptsize{$\pm$2.24} 
    & 87.31\scriptsize{$\pm$0.56} \\
    & GCOPE &   85.01\scriptsize{$\pm$1.00} &84.63\scriptsize{$\pm$1.30} & 85.21\scriptsize{$\pm$0.89} \\
    & \textbf{\texttt{RiemannGFM}}  & \textbf{99.40\scriptsize{$\pm$1.72}} & \textbf{92.75\scriptsize{$\pm$0.61}} & \textbf{93.21\scriptsize{$\pm$0.03} }\\
    \hline
    \multirow{3}{*}{WikiCS}
    & OpenGraph 
    & 89.64\scriptsize{$\pm$3.45} 
    & 72.24\scriptsize{$\pm$4.24} 
    & 86.89\scriptsize{$\pm$3.12} \\
    & GCOPE &   88.51\scriptsize{$\pm$0.47} &89.07 \scriptsize{$\pm$0.58} & 86.09\scriptsize{$\pm$1.14} \\
    & \textbf{\texttt{RiemannGFM}}  &  \textbf{99.31\scriptsize{$\pm$0.02} }&\textbf{92.47\scriptsize{$\pm$0.73}} & \textbf{93.17\scriptsize{$\pm$0.07}} \\
    \hline
    \end{tabular}
        \vspace{-0.12in}
    \end{table}

  \vspace{-0.05in}
\subsubsection{\textbf{Visualization and Discussion}}
\begin{figure}[t]
    \centering
    \begin{subfigure}{0.31\linewidth}
        \centering
        \includegraphics[width=\linewidth]{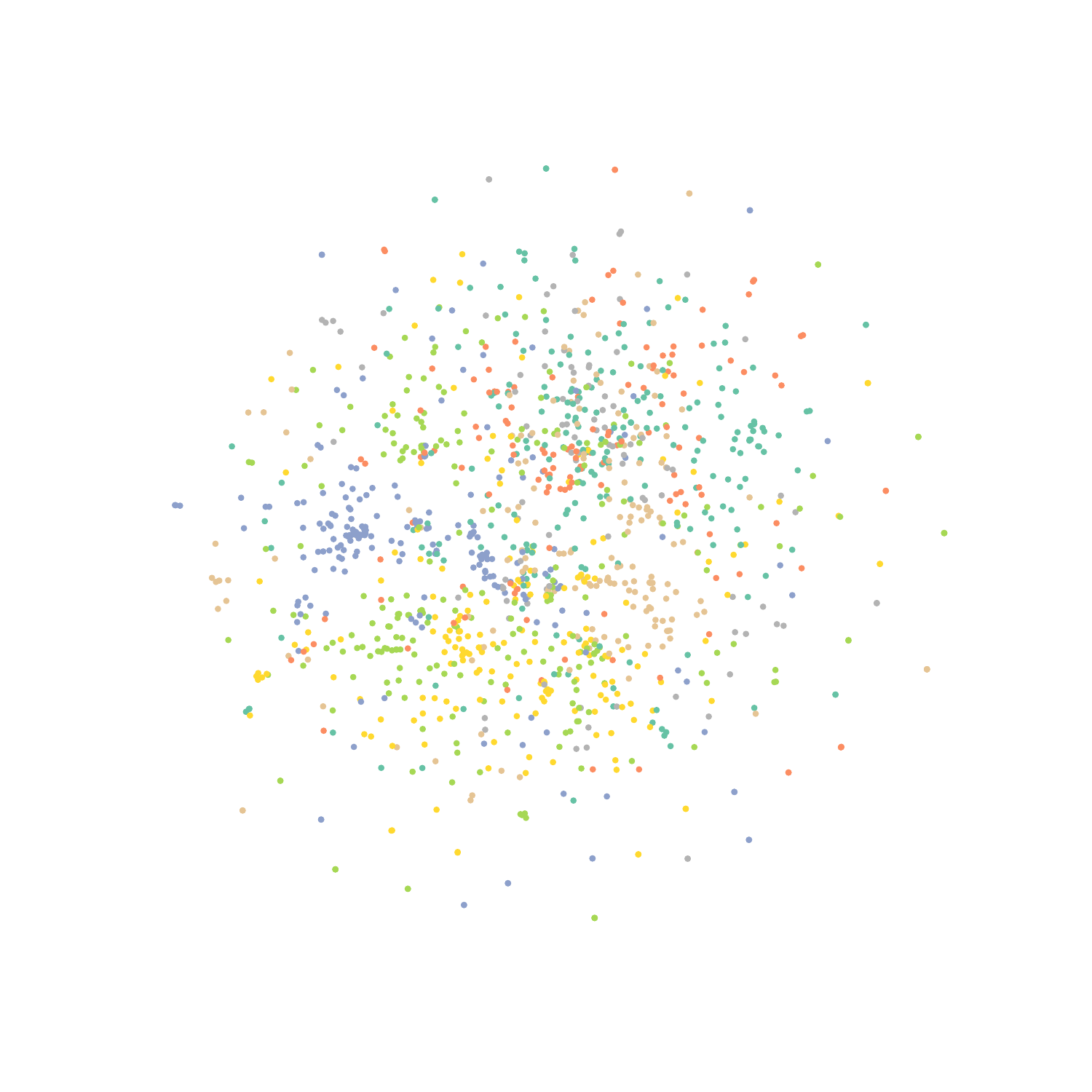}
        \caption{Original}
        \label{fig:Original}
    \end{subfigure}
    \begin{subfigure}{0.31\linewidth}
        \centering
        \includegraphics[width=\linewidth]{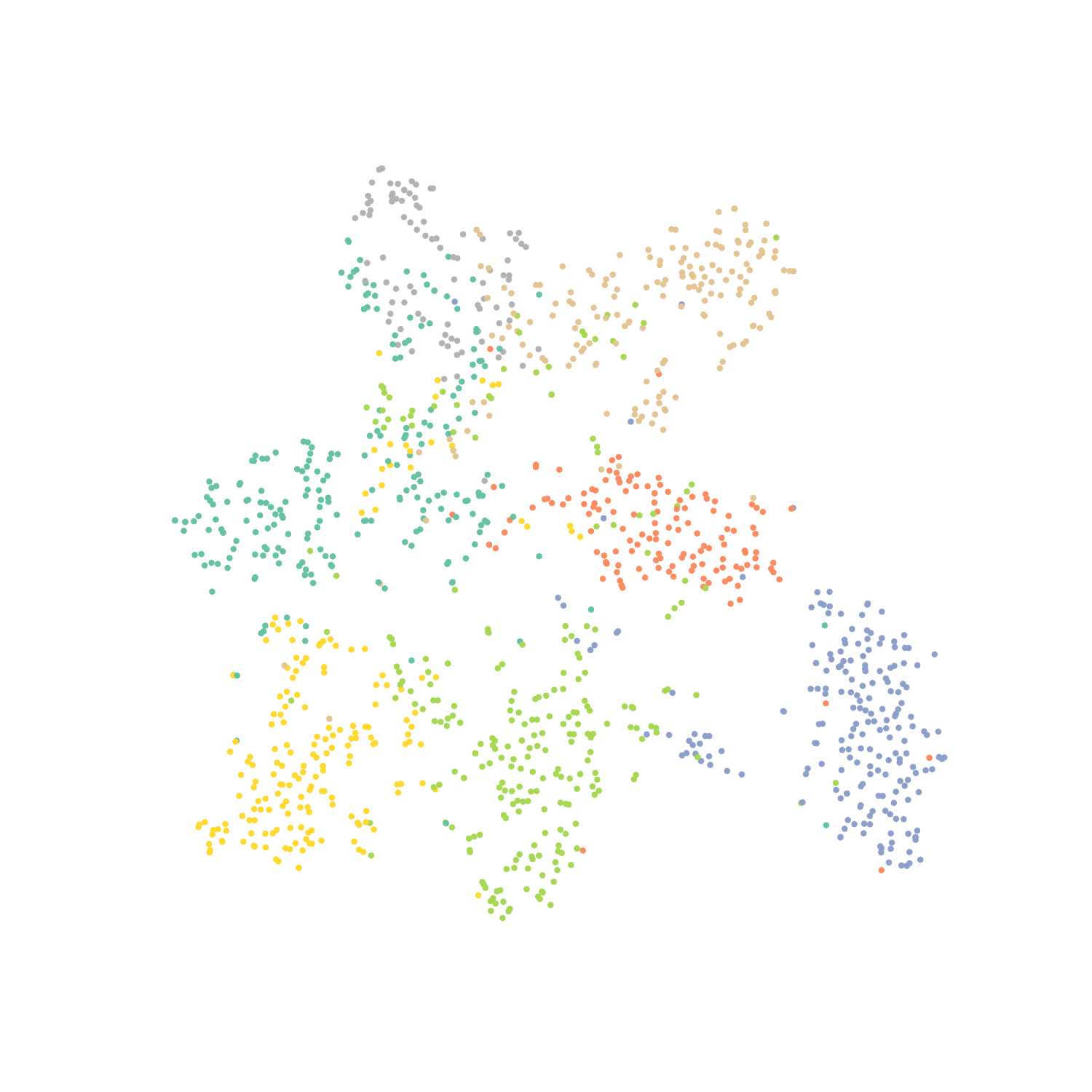}
        \caption{GCN}
        \label{fig:GCN}
    \end{subfigure}
    \begin{subfigure}{0.31\linewidth}
        \centering
        \includegraphics[width=\linewidth]{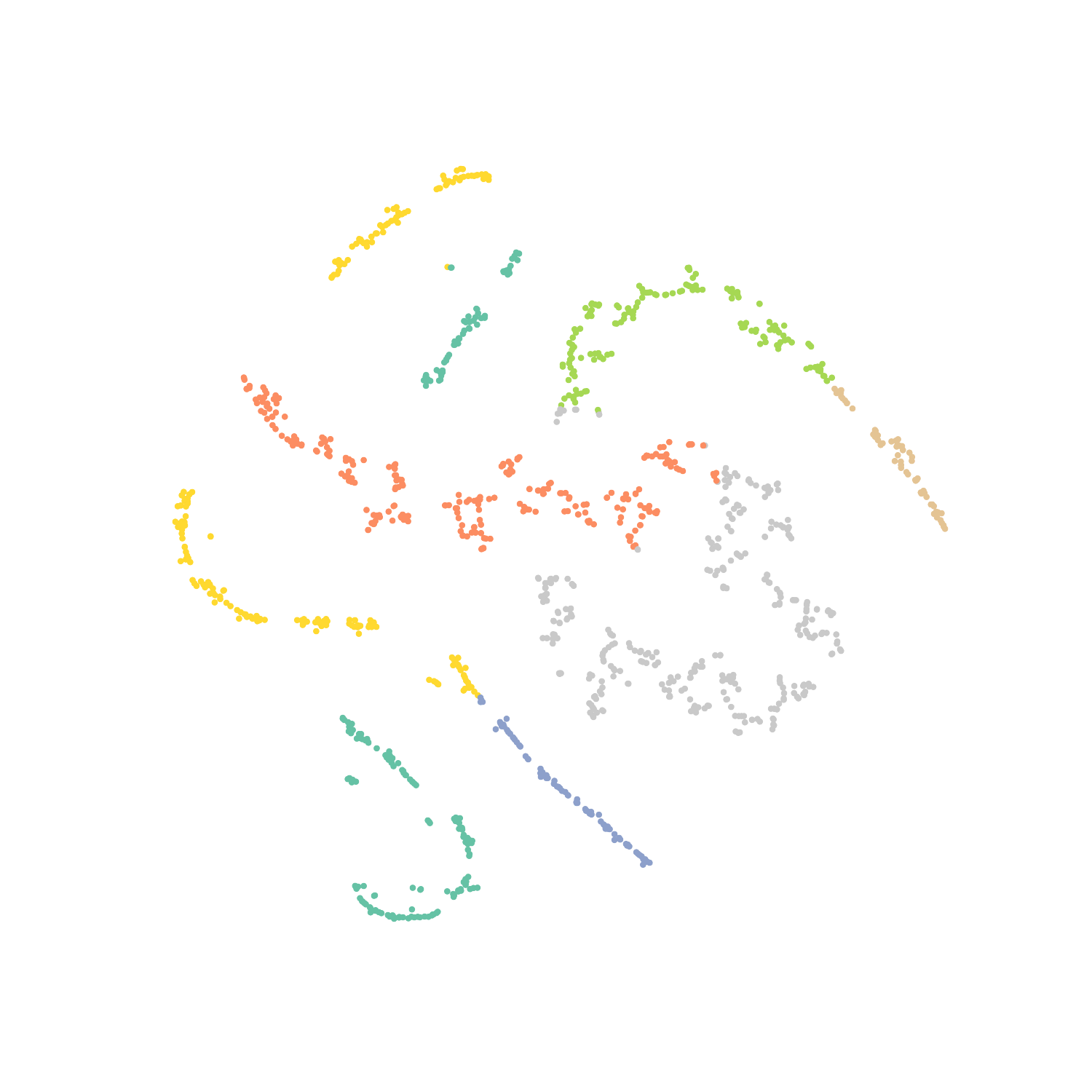}
        \caption{RiemannGFM}
        \label{fig:RiemannGFM}
    \end{subfigure}
        \vspace{-0.1in}  
    \caption{Visualization on Cora}
    \label{fig:tSNE}
    \vspace{-0.15in}
\end{figure}

Here, we visualize the node encoding of Cora via t-sne in Fig. \ref{fig:tSNE}, where different colors denote different node classes.
 Fig. \ref{fig:tSNE}(b) shows the results of GCN, while the visualization of  pre-trained \texttt{RiemannGFM} in  Fig. \ref{fig:tSNE}(c) is given by its node encodings in the shared tangent space of the north pole of Lorentz/Spherical model.
 It shows that the encodings of pre-trained \texttt{RiemannGFM} are more separable than those of a specialized graph model, demonstrating the expressiveness of the knowledge learned in \texttt{RiemannGFM}.
 (\textbf{Additional results are given in Appendix F.})

  \vspace{-0.1in}
\section{Related Work}

  \vspace{-0.03in}
\subsubsection*{\textbf{Graph Neural Network \& Self-supervised Graph Learning.}}
Popular GNNs include graph convolutional nets and graph transformers.
The former conducts neighborhood aggregation with layer-wise message passing \cite{icml19sgc,nips16cnn,nips17GraphSAGE}, while the latter leverages a transformer-like encoder \cite{arxiv22Gransformer}.
Both of them are typically paired with a classification head or reconstructive loss on a specified graph.
Thus, a major shortcoming of traditional graph models is their limited generalization capability.
Self-supervised learning has been integrated into GNNs in recent years \cite{www23graphmae2,iclr19dgi}.
Instead of coupling GNNs with downstream tasks, self-supervised learning conducts parameter training from the graph data itself via specialized pretext tasks. 
However, graph augmentation for self-supervised learning is nontrivial \cite{www2GraphAugmentation,arxiv24GraphContrastiveSurvey}, and the parameters trained on one graph cannot be directly applied to another owing to the difference in attribute distribution.
In other words, existing graph models lack the universality, and are still far from being a foundation model.

  \vspace{-0.05in}
\subsubsection*{\textbf{Graph Foundation Model.}}
% Large language models demonstrate the remarkable success of pre-training a foundation model for multiple downstream tasks \cite{2024gpt4}.
% Recently, efforts have been made to replicate the success to the graph domain.
% \citet{kdd23all-in-one,kdd24gcope} design graph prompting to unify the downstream tasks, corresponding to the prompt of LLM.
% Up to date, the majority of GFMs are designed for text-attributed graphs assisted by the LLM.
% Specifically, LLM-based GFMs are fed with the graph's textual descriptions, e.g., describing nodes/edges throughout the graph \cite{icml24llaga,iclr24ofa}, translating graphs into node sequences \cite{icml24llaga}, and re-framing textual attributes into language \cite{iclr24ofa}.

% Pioneering work \cite{kdd23all-in-one} designs the graph prompting to , the analogy to language prompt.
% GCOPE \cite{kdd24gcope} further conducts model training on multi-domain graphs with coordinators in order to improve the generalization capacity to different datasets.
% Recent efforts have been made to integrate GNN into LLM. 
% For example, LLaGA \cite{icml24llaga} develops a graph translation technique that reshapes a graph into node sequences,
% while OFA \cite{iclr24ofa}  unifies different graph data by describing nodes and edges with natural language.

Recent efforts are generally categorized into two groups.
The first group enhances the vanilla GNNs to achieve better generalization capacity, e.g., unifying the downstream tasks with graph prompt \cite{kdd23all-in-one}, and training on multi-domain graphs with coordinators \cite{kdd24gcope}.
\citet{zhao2024graphany} generalize SGC \cite{icml19sgc} for node classification on any graph.
The second group adapts LLM for analyzing graphs.
LLaGA \cite{icml24llaga} tailors graphs for the language model with node sequences, generated via graph translation,
while OFA \cite{iclr24ofa}  unifies different graph data by the language description of nodes and edges.
OpenGraph \cite{xia2024opengraph} re-frames textual attributes into language with a hierarchy.
Very recently, \citet{xia2024anygraph} propose a mixture of graph experts.
Existing models  typically struggle to maintain the performance on graphs without textual attributes.
Also, they model graphs in Euclidean space, and tend to trivialize the structural complexity.
Distinguishing from the prior studies, we consider graphs in Riemannian geometry, and design the first GFM exploring graph substructures (structural vocabulary).
%, to the best of our knowledge.

% Today's graph foundation models can generally be categorized into two types. 
% The first type is based on traditional Graph Neural Networks (GNNs) or transformer models. [] proposed a task-specific foundation model for graphs across diverse domains. Additionally, [] introduced a graph prompting method, which imitates the prompting style of large language models to facilitate cross-domain graph pre-training.
% The second type leverages large language models (LLMs) to represent graphs through textual descriptions, such as describing nodes and edges separately [] and converting graphs into node sequences []. 

  \vspace{-0.05in}
\subsubsection*{\textbf{Riemannian Manifold \& Graphs.}}
Riemannian manifolds emerge as exciting alternatives for learning graphs \cite{nips24sun,aaai24sun,icdm23sun}.
Concretely, hyperbolic space is well recognized for its alignment with tree-like (hierarchical) structures,
and hyperbolic GNNs show superior results to Euclidean counterparts \cite{nips18hnn,nips19hgcn}. 
The geometric analogy of cycles is hyperspherical space, whose advantage of embedding cyclical structures is reported \cite{22cikmSpherical,nips19sphericalText}. 
\citet{icml20Constant} formulate a graph convolutional net in constant curvature spaces.
Recent advances report the success of Riemannian manifolds in modeling dynamics  \cite{aaai21sun,cikm22sun,aaai23sun,aaai25sun,www23ye} and clustering \cite{icml24sun,www24sun,ijcai23sun}.
We notice that the product manifold has been introduced to study graphs recently, 
and advanced techniques are proposed for node embedding \cite{iclr19mixCurvature,iclr23lantentGraphProduct,aaai22selfMG,cikm24wang}.
However, all of them lack the generalization capability to unseen graph structures, and consider node embeddings on the manifold, while we introduce the notion of tangent bundle to GFM.
%the input graph as a whole and thus lack the generalization capability to other graphs of different sizes.
So far, the potential of Riemannian geometry has not yet been released on GFM, and we are dedicated to bridging this gap.

  \vspace{-0.1in}
  \section{Conclusion}
% Distinguish from seeking a linguistics alignment for GFM, 
% we put forward a fresh perspective of structural geometry that discovers a simple yet effective structural vocabulary and explores the inherent connection between the structural vocabulary and Riemannian geometry.
This work opens a new opportunity to build GFM with a shared structural vocabulary of the graph domain.
Our main contribution is the discovery of tree-cycle vocabulary with the inherent connection to Riemannian geometry, and we present a universal pre-training model \texttt{RemannGFM} accordingly.
Concretely, 
we first propose a novel product bundle to incorporate diverse geometries of the vocabulary. 
On this constructed space, we then stack the Riemannian layers where the structural vocabulary, regardless of specific graphs, is learned on Riemannian manifold.
This offers the shared structural knowledge for cross-domain transferability, 
and informative node encodings  for arbitrary graphs can be generated accordingly. 
Empirical results show the superiority of \texttt{RemannGFM}.
% in cross-domain transfer learning and few-shot learning.

% \begin{table}[t]
%     \vspace{-0.1in}
%     \caption{Comparison of proposed RiemannGFM and previous GFM methods.}
%     \vspace{-0.1in}
%     \resizebox{\linewidth}{!}{ 
%     \begin{tabular}{c   c c  c c }
%     \hline
%              &{\makecell[c]{\textbf{Graph}\\\textbf{Univerisality}}} & {\makecell[c]{\textbf{Translation}\\ \textbf{Free}}} & {\makecell[c]{ \textbf{Task}\\ \textbf{Univerisality}}}    \\
%     \hline
%     One For All \cite{iclr24ofa} &  & &\textcolor{blue}{\ding{51}}   \\
%     LLaGA \cite{icml24llaga} &\textcolor{blue}{\ding{51}}  &  &\textcolor{blue}{\ding{51}}  \\
%     UniGraph \cite{he2024unigraph} &  &  &\textcolor{blue}{\ding{51}} \\
%     OpenGraph \cite{24opengraph} &  &  &\textcolor{blue}{\ding{51}} \\
%     \hline
%     GCOPE \cite{kdd24gcope} &\textcolor{blue}{\ding{51}}  &\textcolor{blue}{\ding{51}}  &\textcolor{blue}{\ding{51}}  \\
%     AnyGraph \cite{xia2024anygraph} &\textcolor{blue}{\ding{51}}  &\textcolor{blue}{\ding{51}}  &\textcolor{blue}{\ding{51}} \\
%     GraphControl \cite{www24graphcontrol} &\textcolor{blue}{\ding{51}}  &\textcolor{blue}{\ding{51}}  & \\
%     GraphAny \cite{zhao2024graphany} &\textcolor{blue}{\ding{51}}  &\textcolor{blue}{\ding{51}}  & \\
%     \hline
%     \textbf{RiemannGFM (Ours)} &\textcolor{blue}{\ding{51}}  &\textcolor{blue}{\ding{51}}  &\textcolor{blue}{\ding{51}} \\
%     \hline
%     \end{tabular} 
%     }
%     \label{compare}
%     \vspace{-0.22in}
%     \end{table}

\begin{acks}
This work is supported in part by NSFC under grants 62202164 and 62322202. Philip S. Yu is supported in part by NSF under grants III-2106758, and POSE-2346158.
\end{acks}


%%

\newpage
\bibliographystyle{ACM-Reference-Format}
\bibliography{www25}

\appendix

%!TEX root = ./main.tex
\section{Notation Table}
\vspace{-0.1in}
\begin{table}[h]
\centering
\caption{Importation Notations.}
\vspace{-0.15in}
\label{table. notation}
\begin{tabular}{c|c}
\hline
\textbf{Notation} & \textbf{Description}   \\
\hline
$\mathcal{M},\mathfrak{g}$        & A smooth manifold and Riemannian metric.        \\
\hline
$\mathcal{T}_x\mathcal{M}$        & The tangent space at $\boldsymbol{x}$.        \\
\hline
$\mathcal{T}\mathcal{M}$         & The tangent bundle surrounding the manifold.        \\
\hline
 $d, \kappa$  &  Dimension and curvature. \\
 \hline
 $\mathcal H$, $\mathcal S$ & Hyperbolic/Hyperspherical space. \\
  \hline
 $\mathcal L$ & A unified formalism of Lorentz/Spherical model.\\
 \hline
 $\boldsymbol{o}$ & North pole of the model space. \\
   \hline
 $\mathcal{G}=(\mathcal{V}, \mathcal{E})$ & A graph with nodes set $\mathcal{V}$ and edges set $\mathcal{E}$.\\
    \hline
 $\boldsymbol{p}\in \mathcal{L}$ & Node coordinate on the manifold.\\
  \hline
   $\boldsymbol{z} \in \mathcal{T}_{\boldsymbol{p}}\mathcal{L}$ & Node encoding in the tangent space.\\
  \hline
 $\phi:\mathcal{L}\times\mathcal{L}\rightarrow \mathbb{R}$ & A parameterized scalar map. \\
 \hline
 $f(\cdot):\mathcal{L}^m \rightarrow \mathcal{L}^n$ & Manifold-reserving linear operation. \\
 \hline
 $[\cdot || \cdot]$ & Vector concatenation. \\
 \hline
 $\operatorname{Exp}_{\boldsymbol{x}}(\cdot)$ & The exponential map at point $\mathbf{z}$ \\
 \hline
 $\operatorname{Log}_{\boldsymbol{x}}(\cdot)$ & The logarithmic map at point $\mathbf{z}$ \\
  \hline
 $\operatorname{PT}_{\boldsymbol x \rightarrow \boldsymbol y}(\cdot)$ & The parallel transport from $\boldsymbol x$ to $\boldsymbol y$\\
 \hline
  % $BC(\cdot)$ & The proposed bundle convolution. \\
  % \hline
\end{tabular}
\vspace{-0.1in}
\end{table}

\vspace{-0.1in}
\section{Proofs and Derivations}

In this section, we detail the proofs of Theorem 1 and 2, and show the derivation of the proposed bundle convolution.

\vspace{-0.03in}
\subsection{The Proposed Linear Operation}
% \newtheorem*{thm6}{Theorem 1 (Manifold-preserving of Proposed Operation)} 
% \begin{thm6}
% Given $\boldsymbol x \in \mathcal L_{\kappa }^{d_1}$ and $\kappa \neq 0$,  
% $f_{\boldsymbol W}(\boldsymbol x) \in \mathcal L_{\kappa }^{d_1}$ preserves on the manifold with  any $\boldsymbol W \in \mathbb R^{{d_1}\times{d_1}}$, and 
% $f_{\boldsymbol W}(\boldsymbol x) \in \mathcal L_{\kappa }^{d_2}$ holds for any $\boldsymbol W \in \mathbb R^{{d_1}\times{d_2}}$.
% \end{thm6}
% \vspace{-0.05in}
\begin{proof}
We give all the key equations, and do not list all the algebra for clarity.
The theorem holds if, with a given curvature $\kappa$, $\kappa \neq 0$, the proposed linear operation satisfies $f_{\boldsymbol W}:\mathcal L_{\kappa }^{d_1}\to \mathcal L_{\kappa }^{d_2}$ for  any $\boldsymbol W \in \mathbb R^{{d_1}\times{d_2}}$.
For $\boldsymbol x \in \mathcal L_{\kappa }^{d_1}$, we conduct the linear operation,
\vspace{-0.03in}
\begin{equation}
f_{\boldsymbol W}(\boldsymbol x)=
\left[\begin{array}{cc}
1 & \mathbf{0}^{\top} \\
\mathbf{0} & \alpha \boldsymbol W
\end{array}\right]
\left[\begin{array}{c}
x_t \\
\boldsymbol x_s
\end{array}\right]
= \left[\begin{array}{c}
x_t \\
\alpha \boldsymbol W\boldsymbol x_s
\end{array}\right].
\vspace{-0.05in}
\end{equation}
With the re-scaling factor $\alpha$ defined as $\frac{\sqrt{\kappa^{-1}-sgn(\kappa)x^2_t}}{\|\boldsymbol W\boldsymbol x_s\|^2}$, the result is yielded as follows
\vspace{-0.1in}
\begin{equation}
f_{\boldsymbol W}(\boldsymbol x)
= \left[\begin{array}{c}
x_t \\
\frac{\sqrt{\kappa^{-1}-sgn(\kappa)x^2_t}}{\|\boldsymbol W\boldsymbol x_s\|^2} \boldsymbol W\boldsymbol x_s
\end{array}\right].
\vspace{-0.05in}
\end{equation}
Given the equality of $sgn(\kappa)x_t^2+{\boldsymbol x}_s^\top{\boldsymbol x}_s=\frac{1}{\kappa}$, it is easy to verify the following equality
\vspace{-0.1in}
\begin{equation}
sgn(\kappa)x_t^2+{\boldsymbol x'}_s^\top{\boldsymbol x'}_s=\frac{1}{\kappa},  \quad
\boldsymbol x'_s=
\frac{\sqrt{\kappa^{-1}-sgn(\kappa)x^2_t}}{\|\boldsymbol W\boldsymbol x_s\|^2} \boldsymbol W\boldsymbol x_s,
\vspace{-0.03in}
\end{equation}
holds for any $\boldsymbol W \in \mathbb R^{{d_1}\times{d_2}}$.
That is, $f_{\boldsymbol W}(\boldsymbol x) \in \mathcal L_{\kappa }^{d_2}$ is ensured, completing the proof.
\end{proof}

\vspace{-0.15in}
\subsection{Geometric Midpoint}
% \newtheorem*{thm7}{Theorem 2 (Arithmetic Mean as Geometric Midpoint)} 
% \begin{thm7}
% %With a set of points and their weights $\{\boldsymbol x_i, \nu_i\}_{i \in \Omega}$, $\boldsymbol x_i \in \mathcal L_{\kappa }^{d}$, $\nu_i \in \mathbb R$,
% The arithmetic mean defined as 
% \vspace{-0.05in}
% \begin{equation}
% mid_\kappa(\{\boldsymbol x_i, \nu_{i}\}_{i \in \Omega})= 
% \frac{1}{\sqrt{|\kappa|} } \sum\nolimits_{i \in \Omega}\frac{\nu_{i} \boldsymbol x_i}{\left| \|\sum\nolimits_{j \in \Omega} \nu_{j} \boldsymbol x_j\|_\kappa \right|}, \ \kappa \neq 0,
% \label{eq:mid2}
% \end{equation}
%  is on the manifold $\boldsymbol c=mid_\kappa(\{\boldsymbol x_i, \nu_{i}\}_{i \in \Omega}) \in \mathcal L_{\kappa }^{d}$,
% and is the geometric midpoint  w.r.t. the squared distance $d$.
% \end{thm7}
\begin{proof}
The theorem claims two facts. The first  is the manifold-preserving of the given arithmetic mean, and the second is the equivalence between the mean and geometric midpoint.
We verify the manifold-preserving by manifold definition $sgn(\kappa)c_t^2+\boldsymbol c_s^\top\boldsymbol c_s=\frac{1}{\kappa}$, for any $\kappa$,
$\kappa \neq 0$.

We elaborate on the geometric midpoint (a.k.a. geometric centroid) before proving the equivalence.
Given the set of points of the manifold $\boldsymbol x_i \in \mathcal L_{\kappa }^{d}$ each attached with a weight $\nu_{i}$, $i \in \Omega$,
the geometric midpoint of squared distance in the manifold $\mathcal L_{\kappa }^{d}$  is given by the following optimization problem,
\vspace{-0.03in}
\begin{equation}
\boldsymbol c=\arg \min\nolimits_{\boldsymbol c \in \mathcal L_{\kappa }^{d}} \sum\nolimits_{i \in \Omega} \nu_{i} d^2_\kappa(\boldsymbol c, \boldsymbol x_i), \ \ \boldsymbol x_i \in \mathcal L_{\kappa }^{d}.
\label{eq:min}
\end{equation}
Now, we derive the geometric midpoint as follows.
Recall the fact that $\langle \boldsymbol{x}, \boldsymbol{x} \rangle_\kappa =\frac{1}{\kappa}$
and
$ d_{\kappa}^{2}\left(\boldsymbol{x}, \boldsymbol{y}\right) = \frac{2}{\kappa} - 2\langle \boldsymbol{x}, \boldsymbol{y} \rangle_\kappa$.
We equivalently transform the minimization of the midpoint in Eq. (\ref{eq:min}) to the maximization  as follows,
\vspace{-0.05in}
\begin{equation}
\boldsymbol c=\arg \max\nolimits_{\boldsymbol c \in \mathcal L_{\kappa }^{d}}   \langle \alpha \sum\nolimits_{i \in \Omega} \nu_{i}\boldsymbol{x}_{j}, \boldsymbol{c}\rangle_\kappa, \\
\end{equation}
where  $\alpha$ is a scaling coefficient so that $\alpha \sum\nolimits_{i \in \Omega} \nu_{i}\boldsymbol{x}_{j} \in \mathcal L_{\kappa }^{d}$ ($\alpha >0$). Note that, for any two points $\boldsymbol{x}, \boldsymbol{y} \in \mathcal L_{\kappa }^{d}$, 
we have the inequality 
$\langle \boldsymbol{x}, \boldsymbol{y} \rangle_\kappa< \frac{1}{\kappa}$
and $\langle \boldsymbol{x}, \boldsymbol{y} \rangle_\kappa = \frac{1}{\kappa}$ if and only if $\boldsymbol{x}=\boldsymbol{y}$.
That is, we need to find an $\alpha$ to satisfy 
$\alpha \sum\nolimits_{i \in \Omega} \nu_{i}\boldsymbol{x}_{j}= \boldsymbol{c}$. 
%Assume 
Let $\alpha_0 >0$ satisfies $\alpha_0 \sum\nolimits_{j \in \hat{\mathcal N}_i} \nu_{ij}\boldsymbol{h}_{j}= \boldsymbol{c}$. 
As the midpoint is required to live in the manifold, i.e., $\alpha_0 \sum\nolimits_{i \in \Omega} \nu_{i}\boldsymbol{x}_{j} \in  \mathcal L_{\kappa }^{d}$, 
we have the following equality
\vspace{-0.03in}
\begin{equation}
\langle \alpha_0 \sum\nolimits_{i \in \Omega} \nu_{i}\boldsymbol{x}_{j}, \alpha_0 \sum\nolimits_{i \in \Omega} \nu_{i}\boldsymbol{x}_{j}\rangle_\kappa = \frac{1}{\kappa},
\vspace{-0.01in}
\label{aggproof}
\end{equation}
according to the definition of the manifold in Eq. (\ref{manifold}), yielding the scaling coefficient as follows,
\vspace{-0.05in}
\begin{equation}
\alpha_0 =  \frac{1}{\sqrt{|\kappa|} \left| ||\sum\nolimits_{i \in \Omega} \nu_{i}\boldsymbol{x}_{j}||_\kappa \right|}>0.
\vspace{-0.03in}
\end{equation}
Consequently, the geometric midpoint is given as
\vspace{-0.03in}
\begin{equation}
mid_\kappa(\{\boldsymbol x_i, \nu_{i}\}_{i \in \Omega})= 
\frac{1}{\sqrt{|\kappa|} } \sum\nolimits_{i \in \Omega}\frac{\nu_{i} \boldsymbol x_i}{\left| \|\sum\nolimits_{j \in \Omega} \nu_{j} \boldsymbol x_j\|_\kappa \right|},
\vspace{-0.05in}
\end{equation}
completing the proof.
\end{proof}

\vspace{-0.1in}
\subsection{Bundle Convolution} 
The unified formalism for Bundle Convolution is given as follows,
\vspace{-0.05in}
\begin{align}
BC_{\boldsymbol p_t}(\{\boldsymbol p_i, \boldsymbol z_i\}_{i\in\Lambda})=\sum\limits_{i\in\Lambda}\left(\alpha_{it} \boldsymbol z_i-\frac{\kappa\alpha_{it}\langle\boldsymbol z_i, \boldsymbol p_t\rangle_\kappa}{1+\kappa\langle\boldsymbol p_i, \boldsymbol p_t\rangle_\kappa}(\boldsymbol p_i+\boldsymbol p_t)\right).
\label{eq:bconv2}
\vspace{-0.05in}
\end{align}
We leverage the equation above to aggregate the node encodings in the corresponding tangent spaces, which span the tangent bundle surrounding the manifold.
The key ingredient of the proposed convolution lies in the parallel transport, which solves the incompatibility issue among different tangent spaces.

The parallel transport w.r.t. the Levi-Civita connection $PT_{x\to y}$ transports a vector in $\boldsymbol v \in \mathcal T_x\mathcal L$ to another tangent space $\mathcal T_y\mathcal L$ with a linear isometry along the geodesic between $\boldsymbol x, \boldsymbol y \in \mathcal L$. 
Concretely, the unit speed geodesic from $\boldsymbol x$ to $\boldsymbol v$ is $\gamma_{\boldsymbol x, \boldsymbol v}(t)=\boldsymbol x\cos_\kappa(t) + \frac{1}{\sqrt{|\kappa|}}\sin_\kappa(t)\boldsymbol v$, for $t\in[0,1]$.
The generic form in $\mathcal L$ is given as
\vspace{-0.05in}
\begin{equation}
PT_{\boldsymbol p_i \to \boldsymbol p_t}(\boldsymbol z_i)=\boldsymbol z_i-\frac{\langle Log_{\boldsymbol p_i}^\kappa(\boldsymbol p_t), \boldsymbol z_i \rangle_{\boldsymbol x}}{d_{\mathcal L}(\boldsymbol p_i,\boldsymbol p_t)}\left(Log_{\boldsymbol p_i}^\kappa(\boldsymbol p_t)+Log_{\boldsymbol p_t}^\kappa(\boldsymbol p_i)\right),
\vspace{-0.03in}
\end{equation}
where $\langle \boldsymbol a, \boldsymbol b \rangle_{\boldsymbol x}=\boldsymbol a^\top\mathfrak{g}_{\boldsymbol x}\boldsymbol b$ is the inner product at the point $\boldsymbol x$, and $\mathfrak{g}_{\boldsymbol x}$

\noindent is  the Riemannian metric of $\mathcal L$ at $\boldsymbol x$.
Given the logarithmic map with curvature-aware cosine  as follows,
\begin{equation}
Log_{\boldsymbol p_i}^\kappa(\boldsymbol p_t)=\frac{\cos^{-1}_\kappa(\beta)}{\sqrt{\beta^2-1}}(\boldsymbol p_t-\beta\boldsymbol p_i), \quad \beta=\kappa\langle\boldsymbol p_i, \boldsymbol p_t\rangle_\kappa.
\end{equation}
The parallel transport in this case is derived as 
\begin{equation}
PT_{\boldsymbol p_i \to \boldsymbol p_t}(\boldsymbol z_i)=\boldsymbol z_i-\frac{\kappa \langle\boldsymbol z_i, \boldsymbol p_t\rangle_\kappa}{1+\kappa\langle\boldsymbol p_i, \boldsymbol p_t\rangle_\kappa}(\boldsymbol p_i+\boldsymbol p_t),  \quad \forall \boldsymbol p_i, \boldsymbol p_t \in \mathcal L,
\end{equation}
where the curvature-aware cosine is defined as $\cos_\kappa(\cdot)=\cos(\cdot)$ when $\kappa>0$, and $\cos_\kappa(\cdot)=\cosh(\cdot)$ with $\kappa>0$, and its superscript $-1$ denotes the inverse function.
Therefore, Eq. (\ref{eq:bconv2}) is given with aggregation over the set $\Lambda$.

\vspace{-0.05in}
\section{Algorithm}

We give the pseudocode of cross-geometry attention in Algo. 2.

\vspace{-0.07in}
\section{Riemannian Geometry}
% A Riemannian manifold $(\mathcal M, \mathfrak g)$ is a smooth manifold $\mathcal M$ endowed with a Riemannian metric $\mathfrak g$. 
% Each point on the manifold is associated with a tangent space where the  metric $\mathfrak g$ is defined.
The curvature is a notion describing the extent of how a manifold derivatives from being ``flat''. 
It is typically viewed as a measure $R(X, Y)Z$ of the extent to which the operator 
$(X,Y) \to \nabla_X \nabla_YZ$ is symmetric, where $\nabla$ is a connection on $\mathcal M$ (where $X, Y, Z$ are vector fields, with $Z$ fixed).
Sectional curvature, defined on two independent vector unit in the tangent space, is often utilized.
The reason is the curvature operator $R$ can be recovered from the sectional curvature, when $\nabla$ is the canonical Levi-Civita connection induced by $\mathfrak g$.
A manifold is said to be a Constant Curvature Space (CCS) if the sectional curvature is constant scalar everywhere on the manifold.

Among Riemannian manifolds, there exist three types of CCSs: the negative curvature hyperbolic space, the positive curvature hyperspherical space, and the zero-curvature Euclidean space.
There are several model spaces of CCSs, e.g., Poincar\'{e} ball model, Poincar\'{e} half-plane, Klein model, Lorentz model, and Stereographical model, and they are equivalent to each other in essence\footnote{They are the same in structure and geometry but have different coordinate systems.}.
In this paper, we opt for the Lorentz/Spherical model\footnote{The Lorentz model of hyperbolic space corresponds to the Spherical model of hyperspherical space in account of the coordinate systems.}, and give a unified formalism,
\vspace{-0.1in}
\begin{equation}
\mathcal L_{\kappa }^{d}=\{ 
\left[\begin{array}{c}
x_t  \\
\boldsymbol x_s 
\end{array}\right] \in \mathbb R^{d+1} 
| \langle \boldsymbol x, \boldsymbol x \rangle_\kappa = \frac{1}{\kappa}, x_t >0, \boldsymbol x_s \in \mathbb R^d\},
\vspace{-0.02in}
\end{equation}
where  $d$ and $\kappa$ denote the dimension and curvature, respectively.
$x_t$ corresponds to the axis of symmetry of the hyperboloid and is termed the time-like dimension, while all other axes $\boldsymbol x_s$ are called space-like dimensions. 
In particular, $\mathcal L^d_\kappa$ becomes the Lorentz model of hyperbolic space under negative $\kappa$, and shifts to Spherical model of hyperspherical space when $\kappa>0$.
Note that, Euclidean space is not included in the formalism, and it requires $\kappa \neq 0$.
The induced hyperbolic space is a $d$-dimensional upper hyperboloid embedded $(d+1)$-dimensional Minkowski space, a.k.a. hyperboloid model. 
Similarly, the corresponding hyperspherical space is also expressed in a $(d+1)$-dimensional space.
All the mathematical construction in this paper is based on the Lorentz/Spherical model.
Accordingly, 
given a point in the manifold $\boldsymbol x \in \mathcal L_{\kappa }^{d}$, the exponential map projects a vector $\boldsymbol  v$ in the tangent space at $\boldsymbol  x$ to the manifold $Exp_{\boldsymbol x}(\boldsymbol v): \mathcal T_{\boldsymbol x}\mathcal L_{\kappa }^{d} \to \mathcal L_{\kappa }^{d}$, and the closed form expression is given as follows,
\vspace{-0.05in}
\begin{equation}
Exp_{\boldsymbol x}(\boldsymbol v) = \cos_{\kappa}(\sqrt{|\kappa|}\|\boldsymbol v\|_\kappa) \boldsymbol x + \sin_\kappa(\sqrt{|\kappa|}\|\boldsymbol v\|_\kappa) \frac{\boldsymbol v}{\sqrt{|\kappa|}\|\boldsymbol v\|_\kappa}. \\
\vspace{-0.03in}
\end{equation}
The logarithmic map $Log_{\boldsymbol x}(\boldsymbol y):  \mathcal L_{\kappa }^{d} \to \mathcal T_{\boldsymbol x}\mathcal L_{\kappa }^{d} $ projects a point $\boldsymbol  y$  in the manifold to the tangent space of $\boldsymbol  x$, serving as the inverse of the exponential map. 
It takes the form of 
\vspace{-0.05in}
\begin{equation}
Log_{\boldsymbol x}(\boldsymbol y) = \frac{\operatorname{cos}_\kappa^{-1}(-\kappa\langle \boldsymbol x, \boldsymbol y\rangle_\kappa)}{\sqrt{\kappa^2\langle \boldsymbol x, \boldsymbol y\rangle_\kappa^2 - 1}} \left(\boldsymbol y + \kappa\langle \boldsymbol x, \boldsymbol y\rangle_\kappa \boldsymbol x\right).
\vspace{-0.03in}
\end{equation}

\begin{algorithm}[t]
    \caption{Cross-geometry Attention in Hyperbolic Space}
    \label{alg. cross-att}
        \KwIn{A substructure, Node coordinates $\boldsymbol p^H$ and $\boldsymbol p^S$, Linear operation $f_{\boldsymbol W}$, A parameterized scalar map $\phi: \mathcal{L}\times\mathcal{L}\rightarrow \mathbb{R}$.}
        \KwOut{The updated node coordinates $\boldsymbol p^H$.}
           Compute the key, query and value via $\boldsymbol k_i=f_{\boldsymbol V}(\boldsymbol p^H_i)$, $\boldsymbol q_i=f_{\boldsymbol Q}(\boldsymbol p^S_i)$ and $\boldsymbol v_i=f_{\boldsymbol V}(\boldsymbol p^H_i)$, respectively;\\
           Compute the score of $\phi([\boldsymbol q_i, \boldsymbol k_j])$ for $i$, $j$ in the substructure;\\
            Derive attentional weight by the softmax of scores over the substructure $\alpha_{ij}=\frac{\exp(\phi([\boldsymbol q_i || \boldsymbol k_j]))}{\sum_{(i,t) \in \Omega}\exp(\phi([\boldsymbol q_i || \boldsymbol k_t]))}$;\\
           Update node coordinate by the weighted geometric midpoint $\boldsymbol v_i=mid_\kappa(\{\boldsymbol v_j, \alpha_{ij}\}_{(i,j) \in \Omega})$;
\end{algorithm}

\vspace{-0.1in}
\section{Experiment Details}

\subsection{Datasets}
We give the statistics in Table \ref{tab:datasets}, and introduce the datasets below.
\begin{itemize}
    \item  \textbf{Citeseer} \cite{citeseerandpubmed} consists of scientific publications in six classes. Nodes and edges denote publications and citation relationship, respectively. Each publication is described as a binary word vector from the dictionary of 3703 unique words.
    \item \textbf{Pubmed} \cite{citeseerandpubmed} is citation network among scientific publications in three classes. Each publication  is described by a TF/IDF weighted word vector from a dictionary  of 500 unique words.
    \item \textbf{GitHub} \cite{github} is a social network where nodes are developers who have starred at least 10 repositories, and edges denote mutual follower relationships. Node features are location, starred repositories, employer, and e-mail address.
    \item \textbf{Airports} \cite{kdd17struc2vec} is a  commercial air transportation network within the United States. The node corresponds to a distinct airport facility, and are stratified into four discrete classes. The edges indicate the existence of commercial flight routes.
    \item  \textbf{ogbn-arxiv} \cite{nips2020arxiv} is the citation network among Computer Science (CS) arXiv papers. Each paper is given as a 128-dimensional feature vector by averaging the embeddings of words in its title and abstract. 
    \item \textbf{Physics} \cite{physics_computers} is co-authorship graphs based on the Microsoft Academic Graph from the KDD Cup 2016 challenge. Nodes and edges denote authors and co-authored relationship, respectively.
    \item \textbf{AComp} (Amazon Computers dataset) \cite{physics_computers} is segments of the Amazon co-purchase graph. Nodes denote goods and edges indicate that two goods are frequently bought.
\end{itemize}

\begin{table}[t]
    \centering
    \caption{Summary of Datasets}
    \vspace{-0.05in}
    \label{tab:datasets}
    \begin{tabular}{l ccc}
    \toprule
    Dataset  & \#(Nodes) & \#(Edges) & 
    Feature Dim. \\
    \midrule
    Cora &   2,708 & 5,429 & 1,433 \\
    Pubmed & 19,717 & 44,338 & 500\\
    GitHub & 37,700 & 578,006& 0 \\
    Airports& 1,190 & 13,599 & 0\\
    ogbn-arxiv& 169,343   & 1,166,243    & 128  \\
    Physics   &  34,493  & 495,924  &  8,415 \\
    AComp& 13,752    & 491,722    & 767 \\
    \bottomrule
    \end{tabular}
     \vspace{-0.1in}
    \end{table}

\vspace{-0.15in}
\subsection{Baselines}
\begin{itemize}
\item \textbf{GCN} \cite{iclr17gcn}  resorts  neighborhood aggregation in spectral domain.
\item \textbf{DGI} \cite{iclr19dgi} introduces a self-supervised paradigm by maximizing the mutual information between the local node view and the global graph view.
\item  \textbf{GraphMAE2} \cite{www23graphmae2} conducts self-supervised learning in the reconstruction of masked node features with masked autoencoders.
\item  \textbf{OFA} \cite{iclr24ofa} describes all nodes and edges with natural language to feed into  LLMs, and  subsequently utilizes  graph prompting  that appends prompting substructures to the input graph.
\item  \textbf{LLaGA} \cite{icml24llaga} re-organizes graph nodes to  sequences and then maps the sequences into the token embedding space via a versatile projector in order to leverage the LLM for graph analysis.
\item  \textbf{OpenGraph} \cite{xia2024opengraph} is  trained on diverse datasets with a unified graph tokenizer, scalable graph transformer, and LLM-enhanced data augmentation, so as to  comprehend the nuances of diverse graphs.
\item  \textbf{GCOPE} \cite{kdd24gcope} is a graph pre-training framework designed to enhance the efficacy of downstream tasks by harnessing collective insights from multiple source domains. 
\item \textbf{GraphAny} \cite{zhao2024graphany} models the inference on a new graph as an analytical solution to a GNN with designs invariant to feature and label permutations and robust to dimension changes.
\end{itemize}

% \paragraph{\textbf{Few-shot settings.}}
% Few-shot learning is a machine learning framework that enables a pre-trained model to generalize to new categories of data that it has not seen during training, using only a few labeled samples for each class.
% This approach, which falls under the meta-learning paradigm (or learning to learn), enables an AI model to make accurate predictions even when trained on a limited amount of labeled data. 
% It is particularly useful in classification tasks where sufficient training data is scarce. 
% In our experiment settings, taking inspiration from previous works \cite{xia2024opengraph}, we retain up to $k$ training instances for labeled classes. This is crucial as it allows the model to learn from a limited number of examples, which is a common scenario in real-world applications where labeled data is scarce. For example, as we pre-train our model on ogbn-Arxiv \cite{nips2020arxiv}, Amazon Computers \cite{physics_computers}, and Coauthor Physics \cite{physics_computers} datasets, then we fetch $k$ samples per class on Citeseer \cite{citeseerandpubmed} and train the classification head. To create the label space for the four datasets, we first load the pre-trained Word2Vec \cite{iclr13word2vec} model like \emph{glove-wiki-gigaword-100}, \footnote{https://radimrehurek.com/gensim/downloader.html} then we map the raw text class names into dense label space. For the classification head, we transform the model output into label space and use cosine similarity to compute prediction scores.
\vspace{-0.11in}
\subsection{Reproducibility \& Implementation Notes}

\vspace{-0.01in}
\subsubsection{\textbf{On Few-shot Learning}}
Few-shot learning performance is significant to evaluate a pre-trained model. In particular, a pre-trained model is examined by classifying new data, which has not been seen during training, with only a few labeled samples for each class.
In our experiment, following the setting of \citet{xia2024opengraph}, we retain up to $k$ training instances for labeled classes.  
For example, we first pre-train our model on ogbn-Arxiv \cite{nips2020arxiv}, Amazon Computers \cite{physics_computers}, and Coauthor Physics \cite{physics_computers} datasets, and then fetch $k$ samples per class on Citeseer \cite{citeseerandpubmed} to train the classification head, so as to infer the classification results.

% Conversely, in link prediction, the k-shot setting translates to maintaining at most k links for each node within the training set. This approach emphasizes the need to make inferences about relationships between nodes with minimal training data, reflecting the challenges faced in dynamic or sparse networks
% For \textbf{zero-shot learning}, we perform tasks without having seen graph structures, node features and node labels during training. Refer to previous works [opengraph,opencitatin], the strategy we used removes class-related parameters. We consider the label classes as separate nodes in our graph and link the vanilla nodes to these class nodes based on their training labels.
% In node classification, the k-shot setting means keeping up to k training instances for each label class, while in link prediction, it involves maintaining at most k links for each node in the training set.
% Non-pre-training methods like GNNs are trained merely on the limited few-shot training data, while other baseline models that use a pre-training-and-tuning paradigm first undergo pre-training before being fine-tuned on the few-shot set.
% In pre-training period, we use the pre-rained model they provide initially. During the subsequent tuning phase, any parameters that cannot be directly transferred between datasets are re-learned. 
% \paragraph{\textbf{Optimizations.}}
% Our model is implemented using Pytorch.The optimization process ..., We use ... loss with ... The learnable parameters of ... are initialized using ...

\vspace{-0.05in}
\subsubsection{\textbf{Initialization and Configurations}}
% The default curvatures of hyperbolic space and hyper-spherical space are $-1$ and $1$. The mini-batch sample method is the same as GraphSAGE \cite{nips17GraphSAGE}. To generate trees by the seed nodes, we apply the BFS algorithm starting at each seed node, the tree edge direction is from children nodes to parent nodes. Then we batch the trees together into one graph for parallel computation. To get initial points on manifolds, there are two ways before using the exponential map.: one is to concatenate $0$ at the first position of the vector; the other one is to project the vector on the tangent space at the original points. For the fine-tuning task heads, we concatenate the raw attributes of the graph with the output of \texttt{RiemannGFM}, then use a simple head like GCN \cite{iclr17gcn} layer to get task-specified outputs.

For model initialization, we first compute the normalized graph Laplacian $\boldsymbol{L}=\boldsymbol{I} - \boldsymbol{D}^{-1/2}\boldsymbol{A}\boldsymbol{D}^{-1/2}$ of the given graph, where $\boldsymbol{A}$ is the adjacency matrix and $\boldsymbol{D}$ is the degree matrix.
Second, we conduct eigenvalue decomposition on $\boldsymbol{L}$  and utilize the largest $K$ eignvectors as node encodings, where $K$ is a predefined number.
Note that, the initialization process indeed normalizes different graphs with the $K$-dimensional encoding $\boldsymbol z$.
Subsequently, we induce  node coordinates via $\boldsymbol p= Exp_{\boldsymbol o}([0 || \boldsymbol z^\top]^\top)$ so that the coordinates are placed on the manifold $\boldsymbol p \in \mathcal L$, where the reference point is the north pole $\boldsymbol o$.
\texttt{RiemannGFM} allows for mini-batch training, and the mini-batch sampling strategy  is the same as that of  SAGE \cite{nips17GraphSAGE}.

\vspace{-0.05in}
\subsubsection{\textbf{Hyperparameters}}

The hyperparameters are  tuned with grid search. 
In particular, we set the dropout rate as $0.1$ to enhance the model robustness, and set learning rate of the pre-training as $0.01$ to balance convergence speed and stability. 
The  dimension of each factor in the product bundle is set as $32$, that is,  we instantiate \texttt{RiemannGFM} on  $\left(\mathcal H^{32}_{-1} \otimes \mathcal T\mathcal H^{32}_{-1}\right) \otimes \left(\mathcal S^{32}_{1} \otimes \mathcal T\mathcal S^{32}_{1}\right)$.
\texttt{RiemannGFM} is implemented with $2$ Riemannian layers.
The parameterized scalar map in cross-geometry attention is a multi-layer perceptions with one hidden layer, whose dimension is set as $256$.
The model is built on PyTorch, and the optimizer is Adam.
%further details are provided in the anonymous link of \url{https://anonymous.4open.science/r/Geo-GFM-1603}.

% In our experiment, we carefully tuned the model's hyperparameters to ensure optimal performance. Specifically, we configured the model with $2$ layers and included bias terms. To enhance the model's robustness, we implemented dropout with a rate of $0.1$. The embedding dimension of each component was set to $32$, while the hidden layer dimension was chosen to be $256$.
% During the pre-training phase, the learning rate was set to $0.01$ to balance convergence speed and stability. 

\begin{table}[t]
\centering % Center the table within the half-column width
\caption{Geometric ablation on Citeseer, Pubmed, and Airport datasets. Node classification results are reported in terms of AUC (\%). The results are given in the form of mean$\pm$std. $\mathcal R^{32}_{0}$ denotes the Euclidean space.}
         \vspace{-0.03in}
\label{tab:geo-ablation-2}
\begin{tabular}{cc|c c c}
\hline
\textbf{Trees} & \textbf{Cycles} & \textbf{Citeseer} & \textbf{Pubmed} & \textbf{Airport} \\
\hline
$\mathcal H^{32}_{-1}$&$\mathcal S^{32}_{1}$
& \textbf{66.38 $\pm$ 0.31}
& \textbf{76.20 $\pm$ 0.79}
& \textbf{55.29 $\pm$ 2.26} \\
$\mathcal H^{32}_{-1}$&$\mathcal R^{32}_{0}$
& 66.26 $\pm$ 1.45 & 73.10 $\pm$ 6.36 & 50.42 $\pm$ 1.48 \\
$\mathcal H^{32}_{-1}$&$\mathcal H^{32}_{-1}$
& 63.37 $\pm$ 1.69 & 72.26 $\pm$ 2.12 & 52.66 $\pm$ 1.46\\
\hline
$\mathcal H^{32}_{-1}$&$\mathcal S^{32}_{1}$
& \textbf{66.38 $\pm$ 0.31}
& \textbf{76.20 $\pm$ 0.79}
& \textbf{55.29 $\pm$ 2.26}\\
$\mathcal R^{32}_{0}$ &$\mathcal S^{32}_{1}$
& 65.52 $\pm$ 1.46 & 71.12 $\pm$ 8.73 & 50.17 $\pm$ 1.26 \\
$\mathcal S^{32}_{1}$ &$\mathcal S^{32}_{1}$
& 64.26 $\pm$ 1.09 & 71.46 $\pm$ 0.72 & 53.72 $\pm$ 0.46 \\
\hline
\end{tabular}
         \vspace{-0.05in}
\end{table}

    \begin{table}[t]
        \caption{Cross-domain node classification prformance on different pre-training datasets.}
        \vspace{-0.03in}
        \label{tab:diff-train-2}
        \centering
        \begin{tabular}{c  c | cc  c  }
        \hline
        & & \multicolumn{3}{c}{\textbf{Testing Datasets}}\\
        \textbf{Pre-training} & \textbf{Method}  & \textbf{Citeseer} & \textbf{Pubmed}  & \textbf{Airport}  \\
        \hline
        \multirow{3}{*}{Flickr}
        & OpenGraph 
        & 63.16\scriptsize{$\pm$4.45} 
        & 60.35\scriptsize{$\pm$5.53} 
        & 43.32\scriptsize{$\pm$2.23} \\
        & GCOPE &64.47\scriptsize{$\pm$2.87} &72.48\scriptsize{$\pm$0.97} & 36.74\scriptsize{$\pm$2.38} \\
        & \textbf{\texttt{RiemannGFM}}  &  \textbf{65.20\scriptsize{$\pm$1.73}} &\textbf{74.04\scriptsize{$\pm$0.53} }& \textbf{46.13\scriptsize{$\pm$2.78} }\\
        \hline
        \multirow{3}{*}{AComp}
        & OpenGraph 
        & 60.24\scriptsize{$\pm$1.25} 
        & 64.45\scriptsize{$\pm$1.24} 
        & 45.02\scriptsize{$\pm$4.25} \\
        & GCOPE &   63.79\scriptsize{$\pm$0.88} &72.80\scriptsize{$\pm$2.14} & 44.19\scriptsize{$\pm$1.53} \\
          & \textbf{\texttt{RiemannGFM}}  &  \textbf{64.80\scriptsize{$\pm$1.96}} &\textbf{77.00\scriptsize{$\pm$0.42}} & \textbf{49.41\scriptsize{$\pm$1.77} }\\
        \hline
        \multirow{3}{*}{WikiCS}
        & OpenGraph 
        & \textbf{67.54\scriptsize{$\pm$2.24}} 
        & 74.98\scriptsize{$\pm$3.25} 
        & 48.92\scriptsize{$\pm$1.22} \\
        & GCOPE &   65.47\scriptsize{$\pm$2.87} &75.38\scriptsize{$\pm$0.83} & 46.05\scriptsize{$\pm$2.51} \\
        & \textbf{\texttt{RiemannGFM}}  &  66.56\scriptsize{$\pm$1.15}&\textbf{75.78\scriptsize{$\pm$1.36}} & \textbf{51.25\scriptsize{$\pm$1.76}} \\
        \hline
        \end{tabular}
                \vspace{-0.05in}
        \end{table}

\section{Additional Results}
We show the additional results of the geometric ablation in Table \ref{tab:geo-ablation-2} and the impact of pre-training datasets in Table  \ref{tab:diff-train-2}.
The geometric ablation in node classification exhibits the similar pattern to that in link prediction, showing the alignment between trees and hyperbolic space  (and between cycles and hyperspherical space).
As shown in in Table \ref{tab:diff-train-2}, the stable performance of our model demonstrates the superiority of exploring GFM with structural vocabulary (i.e., the common substructures of trees and cycles underlying the graph domain).

\end{document}